\documentclass[journal=jacsat,manuscript=article]{achemso}

\usepackage[T1]{fontenc}
\usepackage[version=3]{mhchem} 
\usepackage{xcolor}
\usepackage{amsmath}
\usepackage{mathtools}
\usepackage{graphicx}
\usepackage{multirow}
\usepackage{makecell}
\usepackage{hyperref}
\usepackage{threeparttable}
\usepackage{soul}
\usepackage{svg}
\usepackage{epstopdf}
\graphicspath{ {./images/} }

\usepackage{xr}
\makeatletter
\newcommand*{\addFileDependency}[1]{
    \typeout{(#1)}
    \@addtofilelist{#1}
    \IfFileExists{#1}{}{\typeout{No file #1.}}
}
\makeatother

\newcommand*{\myexternaldocument}[1]{
    \externaldocument{#1}
    \addFileDependency{#1.tex}
    \addFileDependency{#1.aux}
}

\myexternaldocument{02.tex}

\author{Vishnu Jayaprakash}
\affiliation{Department of Chemical and Materials Engineering, \\University of Alberta, Alberta T6G 1H9, Canada}
\email{jayapra1@ualberta.ca}
\author{Jae Bem You}
\affiliation{Department of Chemical Engineering, \\Kyungpook National University, Daegu 41566, Republic of Korea}
\author{Chiranjeevi Kanike}
\affiliation{Department of Chemical and Materials Engineering, \\University of Alberta, Alberta T6G 1H9, Canada}
\author{Jinfeng Liu}
\affiliation{Department of Chemical and Materials Engineering, \\University of Alberta, Alberta T6G 1H9, Canada}
\author{Christopher McCallum}
\affiliation{Independent Scholar, \\Monona, Wisconsin, The United States of America}
\author{Xuehua Zhang}
\affiliation{Department of Chemical and Materials Engineering, \\University of Alberta, Alberta T6G 1H9, Canada}
\email{xuehua.zhang@ualberta.ca}

\title{Determination of Trace Organic Contaminant Concentration via Machine Classification of Surface-Enhanced Raman Spectra}

\abbreviations{}

\usepackage[acronym]{glossaries}

\makeglossaries

\newacronym{pop}{POPs}{persistent organic pollutants}
\newacronym{sers}{SERS}{surface-enhanced Raman spectroscopy}
\newacronym{r6g}{R6G}{Rhodamine 6g}
\newacronym{agnp}{AgNPs}{Silver Nanoparticles}
\newacronym{ve}{VE}{Vitamin E}
\newacronym{fft}{FFT}{Fast Fourier Transform}
\newacronym{wht}{WHT}{Walsh Hadamard Transform}
\newacronym{rfc}{RFC}{Random Forest Classifier}
\newacronym{cv}{CV}{cross-validation}
\newacronym{knn}{k-NN}{k-Nearest Neighbors}
\newacronym{svc}{SVC}{Support Vector Classification}
\newacronym{cnn}{CNN}{Convolution Neural Network}
\newacronym{pca}{PCA}{Principal Component Analysis}

\begin{document}

\begin{tocentry}
\includegraphics{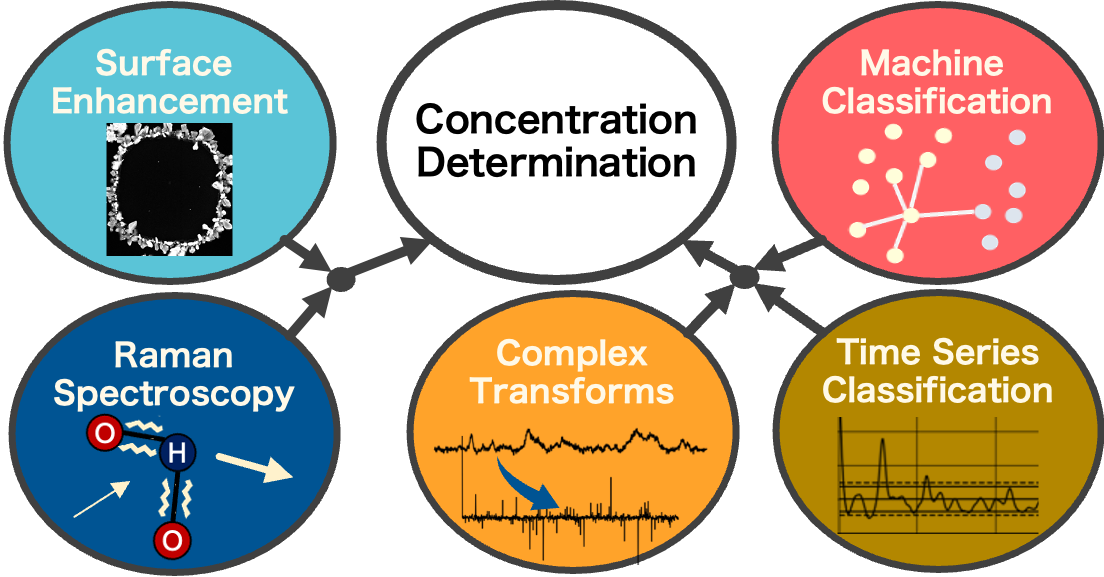}
\end{tocentry}
\pagebreak
\begin{abstract}
Accurate detection and analysis of traces of persistent organic pollutants in water is important in many areas, including environmental monitoring and food quality control, due to their long environmental stability and potential bioaccumulation. While conventional analysis of organic pollutants requires expensive equipment, surface-enhanced Raman spectroscopy (\acrshort{sers}) has demonstrated great potential for accurate detection of these contaminants. However, SERS’s analytical difficulties, such as spectral preprocessing, denoising, and substrate-based spectral variation, have hindered widespread use of the technique. Here, we demonstrate an approach for predicting the concentration of sample pollutants from messy, unprocessed Raman data using machine learning. Frequency domain transform methods, including the Fourier and Walsh-Hadamard transforms, are applied to sets of Raman spectra of three model micropollutants in water (rhodamine 6G, chlorpyrifos, and triclosan), which are then used to train machine learning algorithms. Using standard machine learning models, the concentration of sample pollutants are predicted with $>$80$\%$ cross-validation accuracy from raw Raman data. cross-validation accuracy of 85\% was achieved using deep learning for a moderately sized dataset ($\sim$100 spectra), and 70-80\% cross-validation accuracy was achieved even for very small datasets ($\sim$50 spectra). Additionally, standard models were shown to accurately identify characteristic peaks via analysis of their importance scores. The approach shown here has the potential to be applied to facilitate accurate detection and analysis of persistent organic pollutants by surface-enhanced Raman spectroscopy.
\textbf{Synopsis:} Accurate point of emission detection of water contaminants is limited by the extremely low concentrations, leading to potential build up. This study investigates application of surface-enhanced Raman spectroscopy with machine learning for accurate determination of contaminant levels at low concentrations.

\end{abstract}

\pagebreak
\section{Introduction}
In recent years, there has been a growing concern of the long-term effects of water contamination by persistent organic pollutants (\acrshort{pop}), compounds that are not naturally eliminated by biological systems and can infiltrate many aspects of the ecosystem\cite{EPA2013}. The \acrshort{pop} may include bioactive additives to common consumer, pharmaceutical, and industrial products\cite{StuLap2008}. Many of these \acrshort{pop} are released into the environment at very low concentrations that are difficult to detect. Despite their low concentrations, the long term stability of these compounds potentially leads to bioaccumulation and further spread\cite{Mackay2000}. Compounding on this, many \acrshort{pop} are relatively new compounds whose long-term impacts on the environment and human health have not yet been well defined. Various sectors of human activity produce potentially dangerous \acrshort{pop}: paraben class compounds from consumer cosmetics\cite{LiaLee2013}, bioactive drugs such as acetaminophen and caffeine\cite{DafCac2018}, pesticides such as dichlorodiphenyltrichloroethane and their degradates\cite{EPA2013}, and industrial processing chemicals like polychlorinated biphenyls\cite{Stockholm2001}. In addition to potential bioaccumulation in the environment, recent research has demonstrated the negative impact that these chemicals can have on human reproductive health\cite{RozRed2002}, health of flora and fauna\cite{RevofEnvCont1995}, the human endocrine system\cite{DanHon2011}, and cancer risk\cite{LiQu2011}.

Conventional detection and analysis of \acrlong{pop} involves sensitive chemical analysis techniques such as high performance liquid chromatography (HPLC) or gas chromatography-mass spectroscopy (GC-MS) \cite{JonVoo1999}. However, despite their excellent performance, these techniques are associated with high equipment costs and specialized sample preparation. Additionally, these techniques cannot be performed in-field, as they require a full chemical laboratory. Recently, surface-enhanced Raman spectroscopy (\acrshort{sers}) has been introduced for the detection of \acrshort{pop}\cite{BodSan2020}. \acrshort{sers} is a highly sensitive method that enables both the identification and quantification of target analytes from a sample. For instance, using silver nanoparticles as a \acrshort{sers}-active substrate, Tang et al. demonstrated the detection of 4-mercaptopyridine with concentrations as low as 10$^{-15}$ M\cite{Tang2019}. By combining surface nanodroplet-based nanoextraction and silver nanoparticles, Li et al. showed the detection of pollutants such as methylene blue and malachite green at concentrations near 10$^{-10}$ M.\cite{Li2019} With the existence of handheld and benchtop Raman spectrometers, \acrshort{sers} could prove to be a viable in-field quantification technique. However, the interpretation of \acrshort{sers} data is often very difficult as the intensity and spectral profiles of molecules in \acrshort{sers} are greatly influenced by the orientation with respect to the \acrshort{sers}-active surfaces.\cite{LusThi2020} Moreover, the extensive vibration fingerprints obtained by \acrshort{sers} requires advanced data processing methods such as linear regression or multivariate data analysis for recognition of important features in the data.\cite{LusThi2020} Also, environmental samples may contain many more compounds than the analyte of interest which would lead to complex spectra, making accurate analysis very challenging as peak deconvolution would be required.

The development of machine learning algorithms has enabled the processing of data that had been otherwise impractical. With respect to \acrshort{sers} in particular, machine learning methods have been very useful in analyzing the vibrational fingerprint of molecules from Raman spectra.\cite{Hu2019}  While machine learning-driven Raman analysis has mostly been used for biological and medical applications,\cite{Doherty2021,Lussier2019,Ralbovsky2020} some works have leveraged the advantages of machine learning for chemical analysis. For instance, Zhao et al. developed a machine learning algorithm able to classify the type of edible oil with an accuracy of 96.7\%. The algorithm was trained with Raman data of ten different commercial edible oils.\cite{Zhao2022} Carey et al. showed the identification of minerals by developing full-spectrum matching algorithms based on Raman data.\cite{Carey2015} Detection of pesticide residues in tea using deep learning coupled with \acrshort{sers} was also recently reported.\cite{Zhu2021}

While these advancements have been significant, machine learning has primarily been used for to identify compounds, especially in mixtures\cite{HoJea2021, Carey2015, Zhao2022, Tang2019}. \acrshort{sers} is capable of accurate concentration detection of compounds at even extremely low concentrations, as shown by Li et al. and Tang et al.\cite{LiQu2011}\cite{Tang2019} Despite this, it has been a challenge to employ \acrshort{sers} for concentration determination. Firstly, spectra of the same compound at different concentrations may have minimal correlation. Also, spectra from different concentrations may be indistinguishable from each other due to noise\cite{DabSon2022}, particularly for low concentrations ($<10^{-6}$ M). Finally, many factors may obscure the relationship between peak intensity and the concentration: surface roughness\cite{Kru1994}, surface uniformity\cite{StiDie2008}, laser intensity\cite{JonHoo2019}, and others. Proper implementation of machine learning techniques may be able to reduce the issues posed by these phenomena.

One caveat in applying machine learning methods to \acrshort{sers}, or any other spectroscopic data in general, is the necessity of data preprocessing such as cosmic ray removal, baseline correction and smoothing prior to usage.\cite{Liu2017,Zhu2021,Carey2015} As no standardized method exists, the preprocessing may vary from person to person and application to application, which influences the analysis of data and the generalizability of results. In particular, for low concentrations in which noise is significant, data preprocessing becomes further complicated. Furthermore, the preprocessing strategy used is case specific and must be created based on training data. Overfitting to the training data is one of the largest issues for current machine learning algorithms\cite{Liu2017,WuWan2021}, which is worsened by creating a preprocessing strategy around the training set. Convolutional neural networks (\acrshort{cnn}) have been used for raw, unprocessed spectra in the work of Liu et al., if a sufficiently large dataset is available (>1500 spectra).\cite{Liu2017}

In this work, we demonstrate machine learning strategies to determine the concentration of sample organic pollutants from their Raman spectral data. In particular, our approach is able to assign concentration to the nearest order of magnitude from raw \acrshort{sers} spectra. We use both conventional machine learning algorithms and a \acrshort{cnn} which are trained for unprocessed spectral data. The Fourier and Hadamard Transforms are used to improve the resilience of the model to noise, baseline inclusion, and cosmic rays. Machine learning algorithms such as random forest, support vector classification, k-nearest neighbor, and \acrshort{cnn} were all used in combination with these transforms to create a viable approach to concentration measurement via \acrshort{sers}. Additionally, to address the effect of surface properties and other such interference, this work uses data from two different droplet generation techniques for training. We show that by transforming the raw \acrshort{sers} data, it is possible to develop conventional machine learning algorithms with a cross validated concentration prediction accuracy of $>$0.80 even on noisy, uncorrected datasets from varying data sources. We also demonstrate that with a carefully selected data augmentation procedure and a time series approach for a 1D \acrshort{cnn}, cross-validation accuracy of $>$0.85 is achievable on a sufficiently large dataset with decent quality. The strategy developed in this work expands the application of machine learning-assisted \acrshort{sers} to concentration measurement and allows for the usage of data with minimal preprocessing, low sample size, and high variance due to noise.

\pagebreak
\section{Materials and Methods}
\subsection{Chemical and Materials}
Chemicals were selected for their compatibility with SERS and their relevance as municipal water pollutants: rhodamine 6G, triclosan, and chlorpyrifos. Rhodamine 6G(\acrshort{r6g}) is a commonly used laser dye compound that is well established in \acrshort{sers} literature.\cite{HilSto1984} Additionally, rhodamine dyes have been used in literature to model drug compounds due to their hydrophobic nature.\cite{IyeEdd2019} Despite having minimal associated environmental concern, it is an adequate model compound for SERS analysis purposes.\cite{DabSon2022, KanWu2023} Triclosan is a broad spectrum antibiotic agent that is used in many household products and is seen as a municipal water pollutant. It had been linked to endocrine disruption in humans, as well as acute toxicity in algae.\cite{DanHon2011} Chlorpyrifos is an organophosphate insecticide that can currently be found at residual concentrations in food and drinking water.\cite{MaLuy2020, USEPA2021} In 2021 the US EPA revoked all tolerances for chlorpyrifos due to not being able to establish safe repeated exposure levels from food and drinking water contamination\cite{USEPA2021}. All chemicals were used as supplied without further purification. Water purified from a purification unit (Millipore Corporation, Boston, MA, USA) was used in all the experiments. Sourcing for chemicals is as follows: Rhodamine 6G (\acrshort{r6g}, Fischer Scientific), triclosan (TCI Chemicals, 98\%), and chlorpyrifos (Sigma Aldrich). 

\begin{figure}[H]
    \centering
    \includegraphics[width=\textwidth]{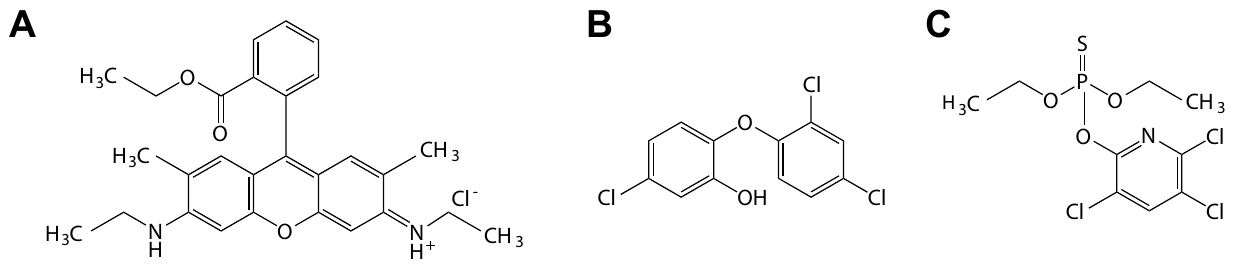}
    \caption{Molecular structures of model compounds explored by the model. (A) \acrshort{r6g}, (B) triclosan, and (C) chlorpyrifos.}
    \label{fig:structures}
\end{figure}

\subsection{Substrate Preparation and Collection of Raman Spectra}

\begin{figure}[H]
    \centering
    \includegraphics[width=\textwidth]{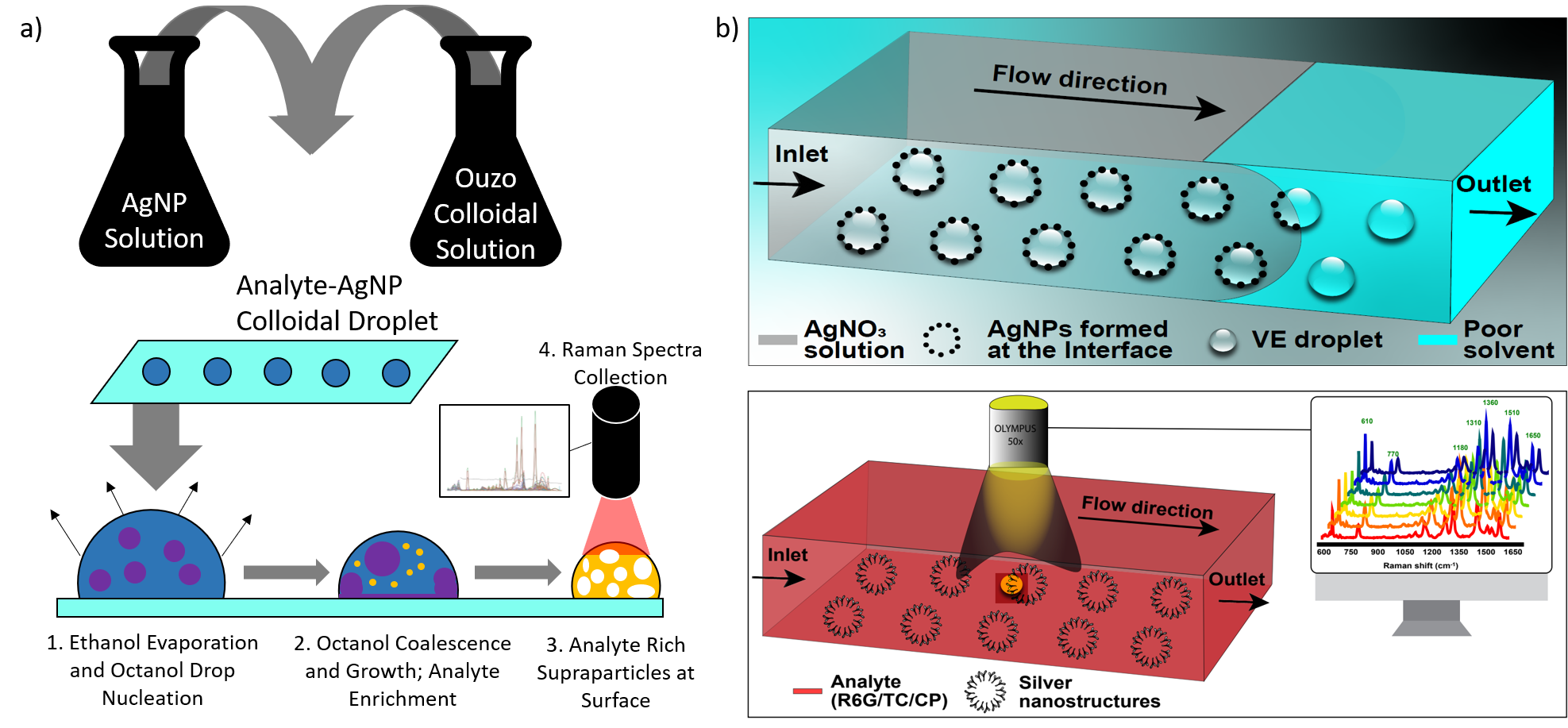}
    \caption{Droplet formation and measurement methods\cite{KanWu2023}. a) Formation of silver-ring nanostructures and method of SERS analysis using microchamber, b) silver supraparticles formed via evaporating Ouzo droplet from colloidal solution.\cite{DabSon2022}}
    \label{fgr:FormDet}
\end{figure}

Raman spectra were obtained for three different environmental compounds: \acrshort{r6g}, triclosan, and chlorpyrifos from other works.\cite{DabSon2022, KanWu2023} Of these chemicals, \acrshort{r6g} and triclosan had two distinct sets of data collected under different conditions. In the first set of \acrshort{r6g} and triclosan data, aqueous samples containing \acrshort{r6g} or triclosan were preconcentrated into a tiny droplet by the evaporating Ouzo method. This was done by forming porous Ag supraparticles using self-lubricating drop evaporation, as shown in Figure \ref{fgr:FormDet}a. Aqueous samples infiltrated with the analyte are formed into a ternary Ouzo solution containing \acrlong{agnp}. Upon evaporation of ethanol from a droplet of Ouzo solution on a substrate, porous Ag supraparticles are formed and the analyte is adsorbed onto the particles, enabling \acrshort{sers} detection.\cite{DabSon2022}

In the second set, all three compounds (\acrshort{r6g}, triclosan, and chlorpyrifos) were tested using Ag nanostructured Si substrate fabricated using a droplet-based approach. Initially, the droplets of vitamin E (\acrshort{ve}) are formed on the hydrophobic microdomains of the patterned Si wafer using a simple solvent exchange method (displacing a good solvent of \acrshort{ve} with a poor solvent). \cite{ChoWei2021} Thereafter, AgNO\textsubscript{3} precursor solution is passed through the microchamber, allowing the reaction to take place at the biphasic interface leading to the nucleation and growth of \acrshort{agnp} as shown in Figure \ref{fgr:FormDet}b. The sample analyte solution is then passed through the microchamber with Ag nanostructured Si wafer for \acrshort{sers} analysis, also in Figure \ref{fgr:FormDet}b.\cite{KanWu2023}

The real-time \textit{in situ} \acrshort{sers} measurements were carried out using Renishaw inVia qontor confocal Raman microscope coupled with an objective of 50$\times$ magnification. Following lasers were used as excitation light source for the detection of model compounds \acrshort{r6g} (633 nm), triclosan (785 nm), and chlorpyrifos (532 nm).  All spectral acquisitions were carried out with 0.1 W power, gratings of 1200 grooves/mm, exposure time of 10 s, and 10 acquisitions with the improved signal-to-noise ratio.\cite{DabSon2022,KanWu2023}

\pagebreak
\subsection{Machine Learning Techniques}
In this work, we employed a classification approach for the prediction of concentration. Raman analysis for the determination of concentration is generally done as order of magnitude (e.g. 10$^{-6}$ M) rather than an exact value. Therefore, instead of treating the concentration as continuous variable, we assigned order of magnitude labels to \acrshort{sers} spectra that act as classes for the machine learning models to sort data into. A classification approach is more in line with existing literature and showed better conditioning from preliminary testing. Literature application of machine learning to \acrshort{sers} data is largely classification based, with a focus on species identification. \cite{LusThi2020,MarHos2020,FanMin2019}

A variety of machine learning models were utilized and compared throughout this study. These include random forest classification (\acrshort{rfc}), k-nearest neighbors (\acrshort{knn}), support vector classification (\acrshort{svc}), as well as a \acrshort{cnn}. These techniques were selected as they are well established classification algorithms that show good general performance across many known problems. Models were compared after tuning via cross-validation (\acrshort{cv}) accuracy. cross-validation is a commonly used machine learning technique that splits training data into sections, using all but one section to train and the remaining section as a validation set, which acts as a practice test. This is repeated, with each section being used as the validation set, so the model that is selected will have the best average performance. This technique limits the model's overfitting to train set and leads to more generalizable results. Spectra were presented with each wavenumber acting as a feature with its respective intensity being the feature value.

For this work an individual treatment was considered to be a chemical dataset (\acrshort{r6g}, triclosan, or chlorpyrifos) or subset with a transform applied (Scaling, Fourier, or Hadamard). The transforms applied to the Raman spectra are explained in the following section. For each treatment, normalization across the samples and scaling across the features were considered as pretreatment options for machine learning. Each treatment was given the pretreatment that produced the best result, with no pretreatment being done if considerable improvements (>3\%) in accuracy could not be made. For these datasets normalization tended to provide inconsistent \acrshort{cv} scores. Scaling always improved performance of algorithms being trained on raw spectral data, but had unpredictable results for the transformed datasets. Scaling of features improved the performance of \acrshort{knn} and \acrshort{svc}, increasing their accuracy to better match that of \acrshort{rfc}, which typically outperformed them.

Hyperparameter tuning was done on all three standard machine learning models via Bayesian search. This technique obtains ideal hyperparameters via a surrogate probability model and gradient descent and is typically the most refined form of hyperparameter tuning. The Bayesian results were validated by a combination of randomized search and grid search, the more simplistic way of determining hyperparameters.

\pagebreak
\subsection{Frequency Domain Transforms}
Prior to using the \acrshort{sers} data for training the algorithm, the raw \acrshort{sers} data was transformed using frequency domain transforms: Fast Fourier (\acrshort{fft}) and Fast Walsh-Hadamard transforms (\acrshort{wht}). Application of \acrshort{fft} and \acrshort{wht} was done to reduce the effect of noise inherent to low concentration data, which machine learning is particularly sensitive to.\cite{MarCom1975} The \acrshort{wht} is a special case of the Fourier transform and is currently used throughout signal processing, filtering and analysis.\cite{Hadamard2011} In contrast with the Fourier transform which outputs both real and imaginary components after the transform, the \acrshort{wht} decomposes a signal into a set of basis functions called Walsh functions with values of +1 or -1. As \acrshort{wht} involves only real components, it is particularly suited to signal processing\cite{Hadamard2011}, and by extension, also suited for machine learning applications.

To perform \acrshort{wht}, the signal (e.g. raw Raman spectra) was multiplied by a Hadamard matrix. A Hadamard matrix is constructed recursively out of previous Hadamard matrices as described by Equation \ref{eqn:HadamardMatrix}. The base unit for this recursion is a 2 $\times$ 2 matrix of ones, with a negative one in the bottom diagonal element. This base unit is then put into a 2 $\times$ 2 matrix to create the next recursion, with each element being the base 2 $\times$ 2 matrix, and the bottom diagonal element being multiplied by negative 1. This can alternatively be represented by the Kronecker product as shown in Equation \ref{eqn:HadamardTransform}. Therefore, Hadamard matrices are always \(2^m \times 2^m\) in size, with \(m\) being selected to create a matrix with dimensions equal to or larger than the signal vector. When the Hadamard matrix is larger than a signal, trailing zeros are appropriately added to the signal vector. The Hadamard Transform when implemented as the Fast Walsh Hadamard Transform is of order $O(n\cdot ln(n))$. This means that it is comparable to \acrshort{fft} in computational complexity.

\begin{equation}
H_1 = 1 \qquad
H_2 = \begin{bmatrix}1 & 1\\1 & -1\end{bmatrix} \qquad 
H_{2^{n}} = \begin{bmatrix}H_{2^{n-1}} & H_{2^{n-1}}\\H_{2^{n-1}} & -H_{2^{n-1}}\end{bmatrix}
\label{eqn:HadamardMatrix}
\end{equation}

\begin{equation}
H_{m} = H_{1}\otimes H_{m-1}, \quad m > 1 \qquad H(f(t)) = H_n \times \Vec{f}, where\ n \geq len(\Vec{f})
\label{eqn:HadamardTransform}
\end{equation}

\begin{figure}[h]
\centering
\includegraphics[scale=0.45]{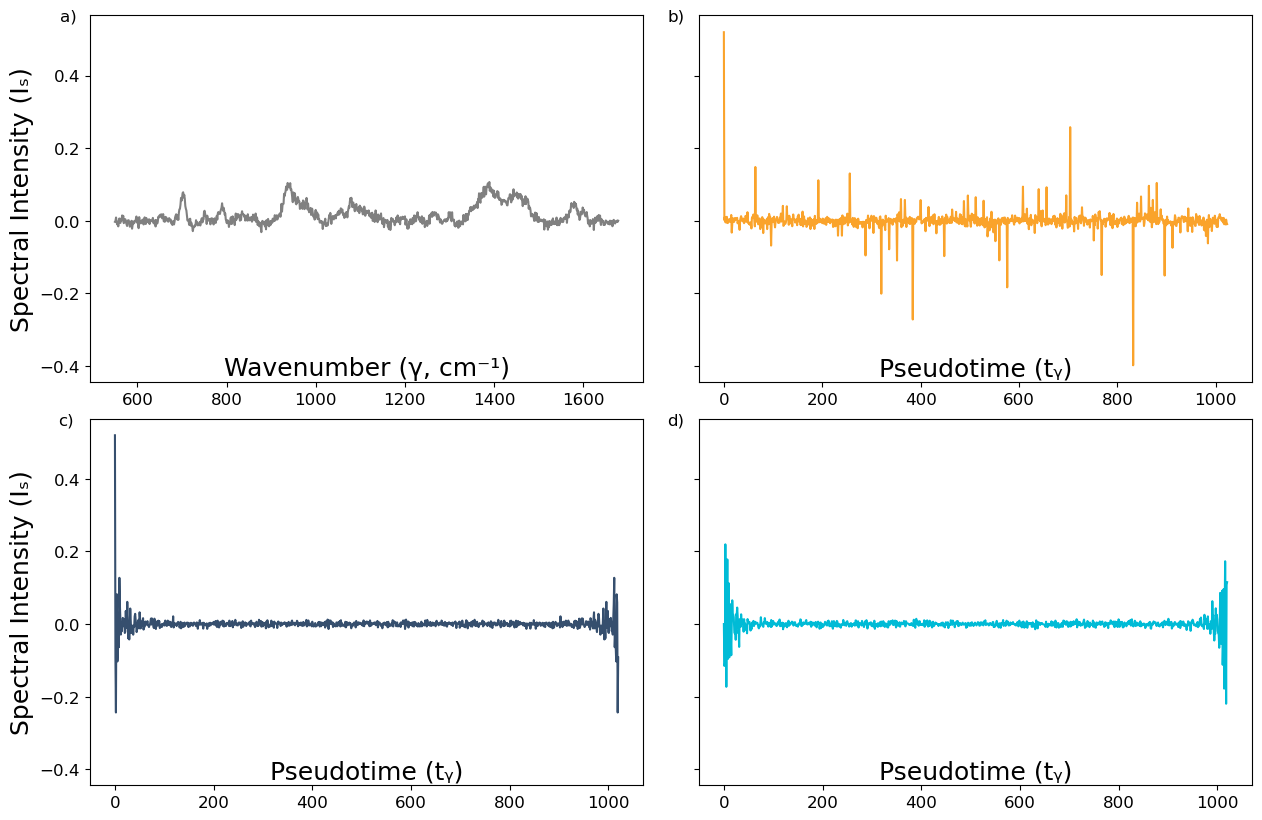}
\caption{Example triclosan Raman spectrum before and after transformation (normalized): a) scaled Raman triclosan spectra, b) Walsh-Hadamard transform of triclosan spectra, c) real part of Fourier transform of triclosan spectra, d) imaginary part of Fourier transform of triclosan spectra.}
\label{fgr:TransformedSpectra}
\end{figure}

Raman spectra are naturally recorded as a frequency domain representation of time domain spectral scattering. Applying a frequency domain transform to frequency spectra will act as a quasi-inverse transform. This places the transformed spectra in some time domain that does not directly match the original time domain that was measured by the instrument. This new time domain is referred to in this work as pseudotime $t_{\gamma}$ and results in transformed spectra being a time series. This change is a functional one, time series data is well explored by machine learning in relation to stock market, weather, and human activity data.\cite{MajLah2020,TaiBak2015,TimeSeriesClass} Figure \ref{fgr:TransformedSpectra} provides a comparison between the original recorded Raman spectra and its respective Hadamard and Fourier transforms.

\pagebreak
\subsection{Hyperparameter Tuning}
Hyperparameter tuning was separately done for all three machine learning methods and for each dataset/subset. Five-fold cross-validation was utilized when tuning hyperparameters, except in the case of triclosan data which could only support three folds due to limited number of spectra available for some concentrations. For the all standard machine learning models, a Bayesian algorithm was used to determine the ideal hyperparameters. To tune the \acrshort{rfc} model, max features, number of estimators, and criterion were varied. To tune the \acrshort{knn}, distance metric, number of neighbors and weights were varied. To tune the \acrshort{svc}, kernel, regularization parameter (C), and polynomial degree were varied.

\begin{table}[!ht]
    \centering
    \begin{threeparttable}
    \begin{tabular}{|l|l|c|c|c|c|c|}
    \hline
        \makecell[c]{\textbf{Learning}\\\textbf{Model}} & \makecell[c]{\textbf{Hyper}-\\\textbf{parameter}} & \makecell[c]{\textbf{R6G}} & \makecell[c]{\textbf{Ouzo}\\\textbf{(R6G)}} & \makecell[c]{\textbf{AgNano}\\\textbf{(R6G)}} & \makecell[c]{\textbf{Triclosan}} & \makecell[c]{\textbf{Chlor}-\\\textbf{pyrifos}} \\ \hline
        \multirow{3}{2.5cm}{\makecell[c]{\textbf{Random}\\\textbf{Forest}\\\textbf{Classifier}}} 
        & \makecell[l]{\textit{n\_estimators}} & 139 & 53 & 166 & 63 & 200 \\
        & \makecell[l]{\textit{max\_features}} & 43 & 12 & 24 & 10 & 148\\
        & \makecell[l]{\textit{criterion}} & Entr. & Gini & Entr. & Entr. & Entr.\\
        \hline
        \multirow{3}{2.5cm}{\makecell[c]{\textbf{k-Nearest}\\\textbf{Neighbors}}} 
        & \makecell[l]{\textit{metric}} & Eucl. & Manh. & Manh. & Mino. & Manh. \\
        & \makecell[l]{\textit{n\_neighbors}} & 2 & 3 & 4 & 2 & 2 \\
        & \makecell[l]{\textit{weights}} & Distance & Uniform & Distance & Uniform & Distance \\
        \hline
        \multirow{3}{2.5cm}{\makecell[c]{\textbf{Support}\\\textbf{Vector}\\\textbf{Classifier}}}
        & \makecell[l]{\textit{C}} & 100 & 100 & 35.748 & 100 & 100 \\
        & \makecell[l]{\textit{degree}} & 6 & 6 & 2 & 2 & 6 \\
        & \makecell[l]{\textit{kernel}} & Linear & Linear & RBF & RBF & Linear \\
        \hline
    \end{tabular}
    \caption{Final hyperparameter settings for scaled R6G dataset, as calculated by Bayesian search. R6G tuned as the full dataset - labelled R6G, the subset of data that uses the evaporating Ouzo droplet technique - labelled Ouzo, and the subset that uses silver nanorings for droplet formation - labelled AgNano. Triclosan and chlorpyrifos were only tuned with the full dataset.}
    \begin{tablenotes}
    \item Entr. - Entropy Criterion, Eucl. - Euclidean Distance, Mino. - Minoski Distance, Manh. - Manhattan Distance, RBF - Radial Basis Function.
    \end{tablenotes}
    \end{threeparttable}
\end{table}

\pagebreak
\subsection{The Convolutional Neural Network}
The Convolutional Neural Network required a design procedure that was independent of the traditional models. Current literature relating machine learning and Raman spectra is focused on classifying species or identifying species in mixtures.\cite{FanMin2019,WuWan2021, HoJea2021} Raman spectra of differing compounds vary in their key peaks, whereas for spectra of the same chemical with different concentrations, peak locations are expected to be very similar with varying intensities. Therefore, the architecture used was based off architecture used to classify time series in the UCI Human Activity Recognition Database.\cite{Bro2020} Spectral data, like time series data, is naturally unstructured and contains features that are connected sequentially to other features.\cite{FawFor2019}

\begin{figure}[H]
\includegraphics[scale = 0.51]{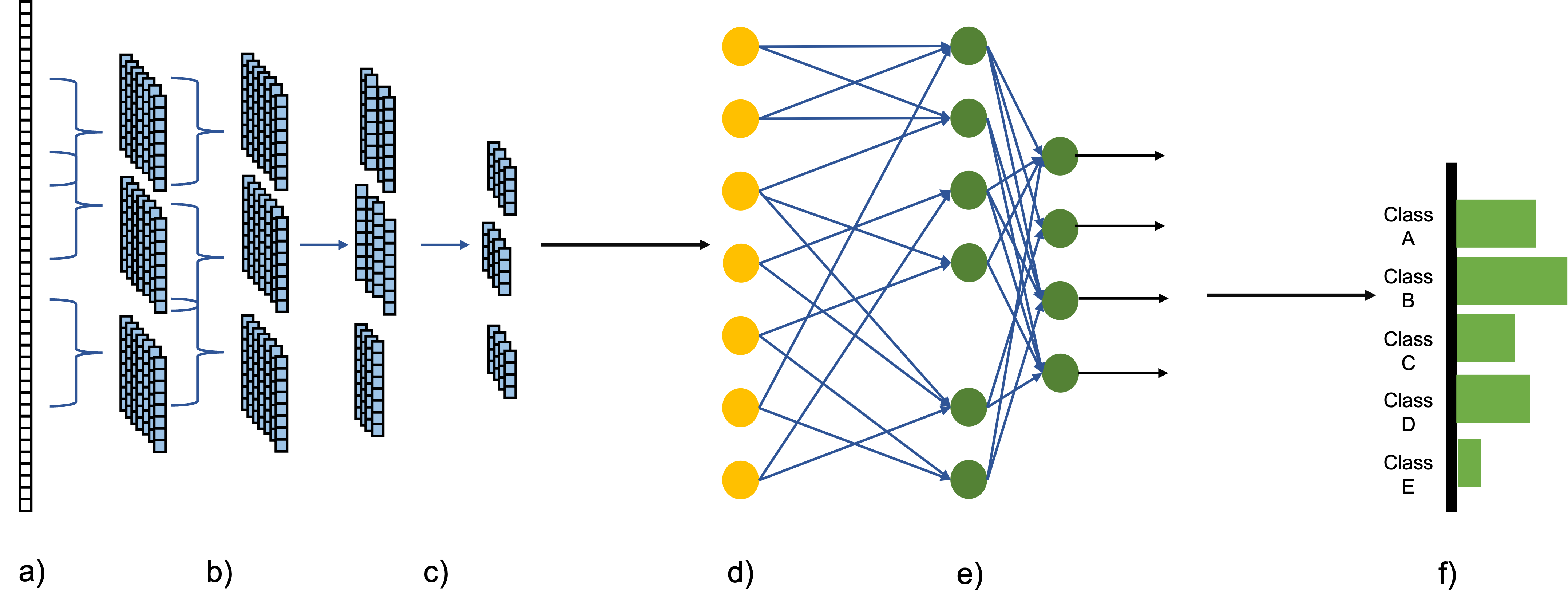}
\caption{Diagram of Convolutional Neural Network architecture. a) Original spectra, b) transformation via two 64 filter 1-D convolution layers with 'relu' activation, c) application of a 50\% dropout layer and a 1-D maxpooling layer, d) flatten layer, e) two fully connected dense layers (100 units 'relu' activation then 8 units 'softmax' activation),  f) probability distribution of classes after softmax, with most probable class selected as answer.}
\label{fgr:CNNDia}
\end{figure}

Figure \ref{fgr:CNNDia} is a schematic of the convolutional neural network architecture used in this work, which consisted of two 1-D convolution layers, one 50\% dropout layer, one maxpooling layer with pool size 2, a flatten layer, and two densely connected layers. Convolutional layers were activated with relu, and dense layers with relu and softmax, respectively. Categorical crossentropy was used as a loss function with an Adam optimizer. All three treatments of data (Scaling, \acrshort{wht}, and \acrshort{fft}) were tested with the \acrshort{cnn}. The dataset was split as 0.81/0.09/0.1 for training/validation/test sets for basic evaluation and generation of learning curves, which are available in the Figure S5-7 of the supporting material. Validation set loss was used as the learning evaluation parameter. Training was done with a batch size of 50 and with 20 epochs. Batch sizes of 5, 20 and 100 as well as epochs of 5 and 50 were also tested with no improvement over reported results. Further optimization of the model architecture would likely improve performance, however was out of the scope of this work. 

Final model comparison results were done via a 5-fold cross-validation with other parameters remaining the same. cross-validation computations were performed on a single node with 20 hyper-threaded cores via the Niagara supercomputer.\cite{LokGru2010,PonZon2019} Overfitting to the training data is a major concern in deep learning applications.\cite{LusThi2020} Using a dataset that integrates multiple droplet formation techniques and spectral collection process improves generalizability of the final model. However, variations in the fundamental spectral character of datasets from differing source, such as interference from droplet media, effects of substrate, and measurement inconsistencies will be a part of the models decision making. An ideal concentration determination model should be trained on a dataset that incorporates even more different data sources so the model's reliance on the specific characteristics of each source is limited.

\pagebreak
\subsection{Data Augmentation}
In order to obtain a dataset large enough for the successful application of \acrshort{cnn}, data augmentation was required. For the generation of new and realistic augmented spectra, existing spectra were randomly selected from the dataset and modified. Modification of the spectra was done by changing three key aspects of the spectral character: offset, peak stretch, and number of single occurrence peaks. These modifications are described in mathematical language in Equations \ref{eqn:Offset}, \ref{eqn:Stretch}, and \ref{eqn:Flip}. Firstly, offset is associated with the inclusion of baseline in datasets without baseline correction, and was not modified for data that had already been baseline corrected. To modify offset, the selected spectra ($\zeta$) had its offset increased or decreased ($\delta O$) by between 0 and 10\% of the standard deviation of all offsets present in the dataset ($\bar O$). 

Next, peak stretch is associated with the natural variation in peak intensity as a result of the orientation of the molecule with respect to the \acrshort{sers} substrate.\cite{LusThi2020} It may also occur as a result of spatial non-uniformity in the \acrshort{sers} substrate. This characteristic is modified by multiplying the intensity values by a stretch factor ($1+\delta S$). This stretch factor is 10\% of the standard deviation of the amplitude variation present in the dataset ($\bar S$). 

Finally, due to the high levels of noise relative to signal for some of these chemicals at low concentrations, single occurrence peaks are observed. The algorithm considers any peak that is within 20\% of the intensity of the largest peak in the spectra to be a significant peak. Single occurrence peaks are falsely significant peaks that occur due to noise. These peaks are present in one spectra of a particular concentration, but not in any others. To represent this in the dataset, a random selection of 0-5 non-significant peaks ($P_{flipped}$) in the selected spectra are taken and stretched by a stretch factor ($x$) which is randomly selected from a log-normal distribution. This has the effect of creating random false significant peaks in the augmented data, which will better represent the noisy data. This is only necessary when noise is a significant issue in the dataset, which was determined by the average fraction of single occurrence peaks peaks across the dataset. When the average fraction of single occurrence peaks is >0.5 then over 50\% of the peaks encountered by the algorithm will be related to noise, and therefore the augmentation strategy must represent this. Fraction of single occurrence peaks is shown in Figure S1.

\begin{figure}[H]
\includegraphics[width = 16cm, height = 7cm]{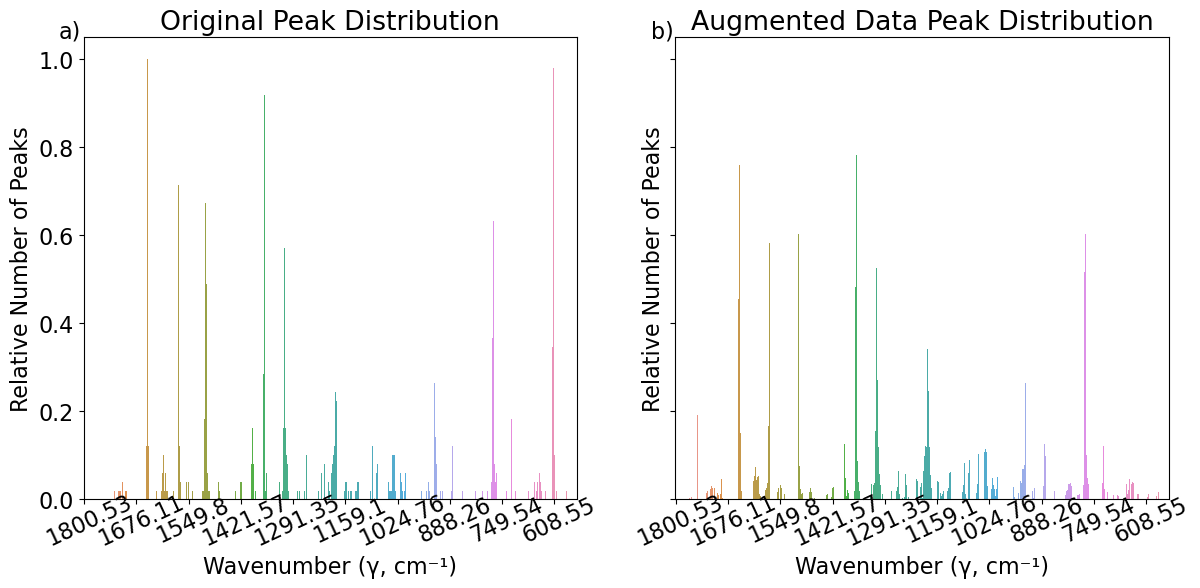}
\caption{Normalized peak distribution across entire R6G dataset (all concentrations). a) Before data augmentation, b) after data augmentation procedure.}
\label{fgr:PeakDistr}
\end{figure}

These modifications of spectral character were selected to match the natural variations of the datasets for a realistic augmentation strategy. The validity of each of the modifications and the extent to which they were done was determined by how they affect the distribution of peaks across the dataset, which is shown in Figure \ref{fgr:PeakDistr}. A representative augmented dataset will have a similar peak distribution to that of the original dataset.

\begin{equation}
\begin{aligned}
\textbf{Offset:     }\delta O\ \in_R\ \{x|\ 0\leq x \leq0.1*\bar O\}\\ 
\bar O = \sigma(\{y\ |\ \forall\ a\in X_{train}, y = min(a)\})\\
\exists\ \delta O \iff  O\ in\ X_{train}\\
\zeta = \zeta + \delta O
\label{eqn:Offset}
\end{aligned}
\end{equation}

\begin{equation}
\begin{aligned}
\textbf{Peak Stretch:     }\delta S\ \in_R\ \{x|-0.1*\bar S \leq x \leq 0.1*\bar S\}\\
\bar S = \sigma(\{y\ |\ \forall\ a\in X_{train},\ y = \frac{max(a)-min(a)}{min(a)}\})\\
\zeta = \zeta * (1 + \delta S)
\label{eqn:Stretch}
\end{aligned}
\end{equation}

\begin{equation}
\begin{aligned}
\textbf{Peak Flip:     } P_{flipped} \subset_R P\ |\ \forall\ p\in P, 0 \leq p \leq 0.2*max(P)\} \\
\exists\ P_{flipped} \iff \frac{|P_{single\ occurrence}|}{|P|} \leq 0.5\ in\ X_{train}\\
P_{new} \equiv P_{flipped} * \{x| x \in_R LogNormal(0, 2) + 1\}\\
\zeta = \{\zeta\ |\ P_{flipped} \subset \zeta = P_{new}\}
\label{eqn:Flip}
\end{aligned}
\end{equation}

\begin{centering}
Where O is an offset, S is a peak stretch factor, P is a set of peaks, p is an individual peak and $\zeta$ is a spectra chosen randomly to be augmented.
\end{centering}

\pagebreak
\section{Results and Discussion}
\subsection{Data Exploration}
In this work, spectra were largely utilized as collected in their original work, which is described in the materials and methods section, with minimal modification. Concentrations in which the number of spectra collected was less than 4 were dropped from the dataset (\(10^{-3}\) M in \acrshort{r6g} evaporating Ouzo and \(10^{-6}\) M in triclosan evaporating Ouzo). Certain spectra required downsampling or truncation to match the length of collected spectra in the other datasets. Downsampling was done to 2.12 $cm^{-1}$, 0.99 $cm^{-1}$, and 1.67 $cm^{-1}$ wavenumber gap for \acrshort{r6g}, triclosan and chlorpyrifos respectively. Truncation was needed for triclosan and chlorpyrifos as the \acrshort{agnp} rings dataset included lower wavenumbers (400 $cm^{-1}$ to 600 $cm^{-1}$) than the Ouzo dataset for these two chemicals. Information loss due to truncation and downsampling is an issue in combining datasets from varying sources.

 The Ouzo droplet method produced results with significantly more noise at low concentrations (<$10^{-7}$ M). Additionally, the first half of the Ouzo data for \acrshort{r6g} was not baseline corrected. The triclosan dataset has a greater variance between spectra collected using the Ouzo method vs. the silver ring method due to larger interference from droplet media. Also, the level of lowest detection for triclosan was reported to be lower than that of \acrshort{r6g} or chlorpyrifos\cite{DabSon2022}. Chlorpyrifos data was obtained only via the silver nanoparticle ring method and had considerable noise, especially at low concentrations. \acrshort{r6g} data was analyzed based on the combined dataset for the chemical as well as individual subsets, while the other chemicals were only analyzed via their combined dataset. Data source and distribution is summarized in Table \ref{tab:SpectraCount}.

\begin{table}[H]
    \centering
    \begin{threeparttable}
    \begin{tabular}{p{1.1cm}|p{1.3cm}|p{1.1cm}|p{1.3cm}|p{1.1cm}|p{1.3cm}|p{1.1cm}|p{1.3cm}|p{1.1cm}|p{1.3cm}}
    \multicolumn{4}{c}{\textbf{R6G}} & \multicolumn{4}{c}{\textbf{Triclosan}} & \multicolumn{2}{c}{\textbf{Chlorpyrifos}}\\
    \hline
    \multicolumn{2}{c}{\makecell[c]{\textit{Evaporating}\\\textit{Ouzo}}} &  \multicolumn{2}{c}{\makecell[c]{\textit{Silver}\\\textit{Nanoparticles}}} & \multicolumn{2}{c}{\makecell[c]{\textit{Evaporating}\\\textit{Ouzo}}} & \multicolumn{2}{c}{\makecell[c]{\textit{Silver}\\\textit{Nanoparticles}}} & \multicolumn{2}{c}{\makecell[c]{\textit{Silver}\\\textit{Nanoparticles}}}\\
    \hline
    \makecell[c]{Conc.\\(M)} & \makecell[l]{Num.\\of\\Spectra} & \makecell[c]{Conc.\\(M)} & \makecell[l]{Num.\\of\\Spectra} & \makecell[c]{Conc.\\(M)} & \makecell[l]{Num.\\of\\Spectra} & \makecell[c]{Conc.\\(M)} & \makecell[l]{Num.\\of\\Spectra} & \makecell[c]{Conc.\\(M)} & \makecell[l]{Num.\\of\\Spectra} \\
    \hline
    \(10^{-5}\) & 18\tnote{*} & \(10^{-5}\) & 10 & \(10^{-5}\) & 6\tnote{**} & \(10^{-3}\) & 5 & \(10^{-3}\) & 10 \\
    \hline
    \(10^{-9}\) & 8\tnote{*} & \(10^{-6}\) & 10 & \(10^{-7}\) & 6\tnote{**} & \(5{\times}10^{-4}\) & 5 & \(10^{-4}\) & 10 \\
    \hline
    \(10^{-11}\) & 14\tnote{*} & \(10^{-7}\) & 10 & \(10^{-8}\) & 6 & \(10^{-4}\) & 5 & \(10^{-5}\) & 10 \\
    \hline
    \(10^{-14}\) & 9\tnote{**} & \(10^{-8}\) & 10 & \(10^{-9}\)  & 6\tnote{**} & \(5{\times}10^{-5}\) & 5\tnote{**} & \(10^{-6}\) & 10\tnote{**} \\
    \hline
    \(10^{-16}\) & 16\tnote{**} & \(10^{-9}\) & 10 & - & - & \(10^{-5}\) & 5\tnote{**} & \(10^{-7}\) & 10\tnote{**} \\
    \hline
    \textbf{Total:} & 65 & & 50 & & 24 & & 25 & & 50
    \end{tabular}
    \caption{Number of spectra at each concentration for all chemicals and subsets.\cite{DabSon2022}\cite{KanWu2023}}
    \label{tab:SpectraCount}
    \begin{tablenotes}
    \item[*] Contains Baseline Uncorrected Data
    \item[**] Contains Increased Noise
    \end{tablenotes}
    \end{threeparttable}
\end{table}

\begin{figure}[H]
\includegraphics[scale=0.5]{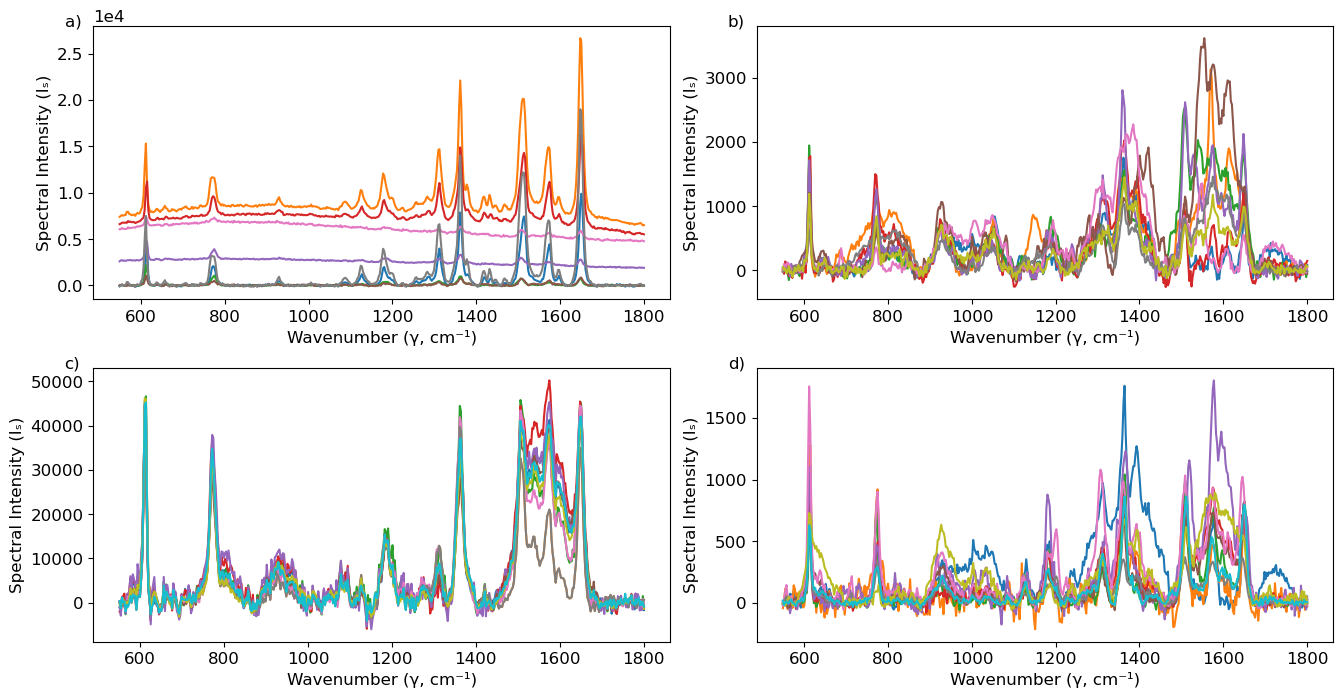}
\caption{Examples of collected R6G spectra and major considerations for machine learning from the dataset. a) Spectra without baseline correction ($[R6G] = 10^{-9}$ M), b) spectra with significant noise ($[R6G] = 10^{-14}$ M), c) clean spectra with baseline correction ($[R6G] = 10^{-6}$ M), d) baseline corrected spectra with high variance ($[R6G] = 10^{-9}$ M)}
\label{fgr:ExampleSpectra}
\end{figure}

Figure \ref{fgr:ExampleSpectra} shows the three major issues in the datasets that machine learning models must learn to handle. Firstly, there are spectra that are not baseline corrected leading to offset, presence of cosmic background rays, and unfiltered noise. Secondly, there are low concentrations where noise from the droplet media creates significant variation in the spectra, even at the key peaks. Finally, due to the high sensitivity of \acrshort{sers} as a measurement technique, non-uniformity of the substrate results in high variance in some parts of the dataset.

\begin{figure}[H]
\includegraphics[trim={1cm 0 0 0}, clip, scale= 0.387]{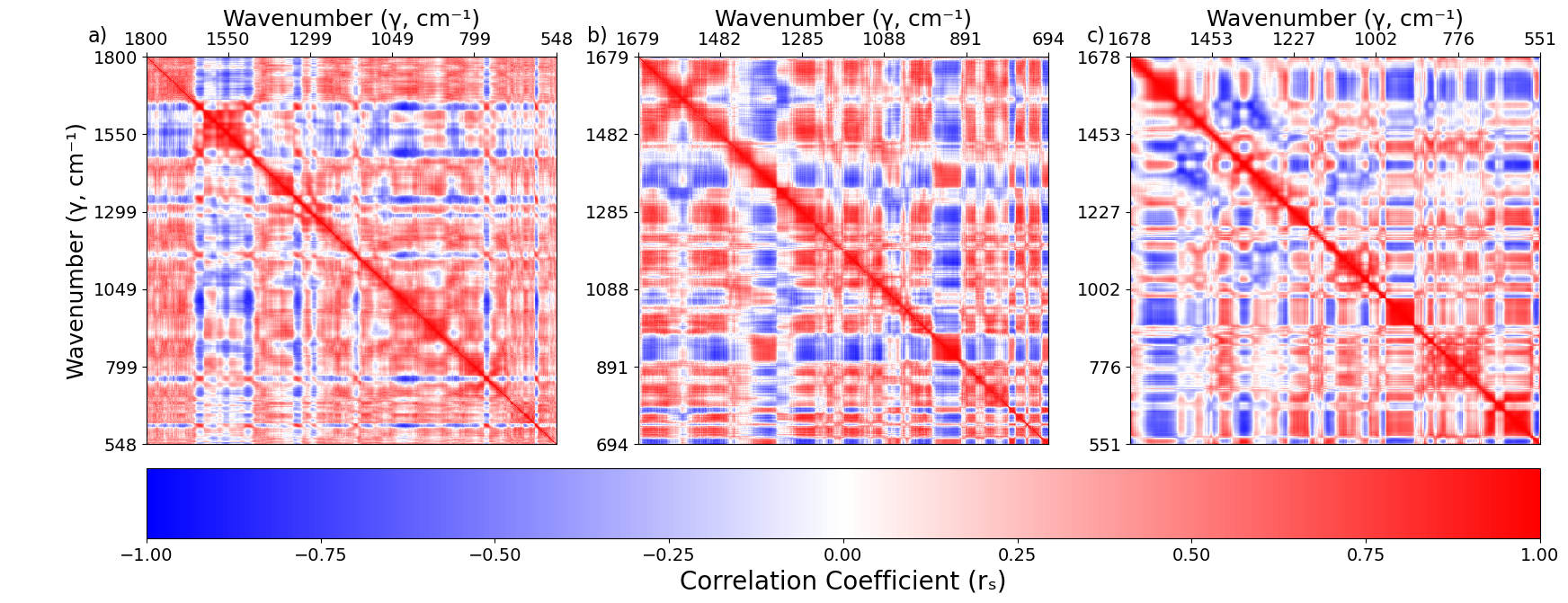}
\caption{Spearman correlation matrices for each chemical (normalized spectra). a) \acrshort{r6g} (average Spearman correlation coefficient = 0.209), b) triclosan (average Spearman correlation coefficient = 0.174), c) chlorpyrifos (average Spearman correlation coefficient = 0.125).}
\label{fgr:CorrMat}
\end{figure}

Figure \ref{fgr:CorrMat} displays correlation matrices for all three chemicals. The Spearman's rank correlation coefficient is used for correlation matrices instead of Pearson's correlation as the relationship is expected to be nonlinear. Spearmans's rank coefficient describes how well a two variable relationship can be described by a monotonic function (strictly increasing or strictly decreasing). A value of 1 or -1 represents a relationship that is strictly increasing or decreasing, respectively. Values nearer to zero indicate a non-monotonic, potentially random, relationship. For the correlation plot of the chemical datasets there are clusters of red along the main diagonal that would correspond to wide peaks which rise monotonically together. Along any row or column from a point on the main diagonal is the monotonic correlation of other wavenumbers to the wavenumber on the main diagonal. Correlation matrices are symmetrical along the main diagonal. Clusters of wavenumbers with a correlation coefficient near 1 or -1 represent peaks that are well correlated to the peak of interest in the main diagonal. 

In the \acrshort{r6g} correlation plot there is a moderate positive correlation between most wavenumbers, except for the region around 1550 $cm^{-1}$, which indicates that this region is not from any bond in the chemical and is rather an interference peak from the droplet formation method.  Similarly, in the triclosan correlation plot, there is a region around 1350 $cm^{-1}$ that is highly correlated to the region around 900 $cm^{-1}$ and vice versa. From the spectra in Figure S3, it is seen that peaks in these region are present at higher concentrations but not at low concentrations or in the bulk, however they appear together. This may mean that they are from some portion of the chemical that is only detectable at higher concentrations. When the overall Spearman correlation is close to 1 or -1 then less of the variance is from noise or other confounding variables such as orientation, substrate variation and measurement technique. The average Spearman correlation of all three species is close to zero with \acrshort{r6g} being highest followed by triclosan and then chlorpyrifos. This fits expectation as the datasets' cleanliness also follows \acrshort{r6g}>triclosan>chlorpyrifos.

\pagebreak
\subsection{Standard Model cross-validation Results}
Five fold cross-validation was the main metric for comparison between treatments in the case of standard machine learning models and its results are presented in Table \ref{tab:CrossValTabl}. While each dataset will have a machine learning method that best fits it, random forest classification has the best overall performance. The Hadamard transform performs best in all datasets but chlorpyrifos. As seen in Figure S4 the chlorpyrifos spectra has a relatively simple structure, with only 2-3 characteristic peaks in the portion scanned, with the most important peaks being quite wide making decomposition into frequency domain more muddled. This may explain the relative over-performance of the simple scaling method. Even so, the Hadamard \acrshort{svc} results for chlorpyrifos are comparably high at 92.5\%. Using the best standard machine learning model in combination with the Hadamard transform, prediction accuracies are high for \acrshort{r6g} and its respective subsets ($\geq$85\%). Accuracy of triclosan predictions is limited by variation in methods used to collect the data however, the Hadamard transform produces fair results of 82\%, a 7.5\% increase over simple scaling. Finally, the chlorpyrifos accuracy is quite high even with its considerable noise, likely due to the data all consistently being collected via silver nanoparticles.

For the purpose of cross-validation accuracy no differentiation was made between small mistakes ($10^{-7}$ M classified as $10^{-9}$ M) and large mistakes ($10^{-7}$ M classified as $10^{-16}$ M). An error function that had greater punishment for larger mistakes would need to be application specific with regards to how large and small errors are weighted and therefore was not included in the analysis. All models tend to only make small magnitude errors (single category), but over many trials and random seeds Hadamard transformed data was observed to only make single category errors, even when scaled and Fourier made slightly larger errors.

\begin{table}[H]
    \centering
    \begin{threeparttable}
    \begin{tabular}{|p{2.8cm}|p{2cm}|ccc|}
    \hline
        \makecell[c]{Dataset} & Transform & Random Forest & k-Nearest Neighbors & Support Vector \\ \hline
        \multirow{3}{2.5cm}{\makecell[c]{R6G\\Combined\tnote{*}}} & None & $0.836 \pm 0.051$ & $0.805\pm 0.080$ & $0.784 \pm 0.072$ \\
        \cline{2-5}
        & Fourier & $0.847 \pm 0.083$ & $0.773 \pm 0.075$ & $0.847 \pm 0.043$ \\ 
        \cline{2-5}
        & Hadamard & $0.837 \pm 0.100$ & $0.783 \pm 0.091$ & $\mathbf{0.847 \pm 0.043}$ \\ \hline
        \multirow{3}{2.5cm}{\makecell[c]{Evaporating\\Ouzo\tnote{*,**}}} & None & $0.791 \pm 0.086$ & $0.771 \pm 0.093$ & $0.747 \pm 0.143$ \\
        \cline{2-5}
        & Fourier & $0.862 \pm 0.103$ & $0.827 \pm 0.121$ & $0.809 \pm 0.055$ \\ 
        \cline{2-5}
        & Hadamard & $\mathbf{0.884 \pm 0.071}$ & $0.867 \pm 0.094$ & $0.849 \pm 0.106$ \\ \hline
        \multirow{3}{2.5cm}{\makecell[c]{Silver\\Nanoparticles}} & None & $0.950 \pm 0.061$ & $0.900 \pm 0.094$ & $0.950 \pm 0.100$ \\
        \cline{2-5}
        & Fourier & $1.00 \pm 0.00$ & $0.975 \pm 0.050$ & $\mathbf{1.00 \pm 0.00}$ \\ 
        \cline{2-5}
        & Hadamard & $1.00 \pm 0.00$ & $0.975 \pm 0.050$ & $1.00 \pm 0.00$\\ \hline
        \multirow{3}{2.5cm}{\makecell[c]{Triclosan\tnote{**}}} & None & $0.747 \pm 0.108$ & $0.739 \pm 0.162$ & $0.725 \pm 0.179$ \\
        \cline{2-5}
        & Fourier & $0.797 \pm 0.067$ & $0.744 \pm 0.045$ & $0.742 \pm 0.060$ \\ \cline{2-5}
        & Hadamard & $\mathbf{0.822 \pm 0.081}$ & $0.772 \pm 0.079$ & $0.772 \pm 0.079$ \\ \hline
        \multirow{3}{2.5cm}{\makecell[c]{Chlorpyrifos\tnote{**}}} & None & $\mathbf{0.975 \pm 0.050}$ & $0.825 \pm 0.170$ & $0.850 \pm 0.094$ \\
        \cline{2-5}
        & Fourier & $0.525 \pm 0.215$ & $0.850 \pm 0.050$ & $0.925 \pm 0.100$ \\ \cline{2-5}
        & Hadamard & $0.800 \pm 0.170$ & $0.875 \pm 0.112$ & $0.925 \pm 0.100$ \\ \hline
    \end{tabular}
    \caption{cross-validation results across datasets and transforms. Best performance in bold (high accuracy>low standard deviation>low fit time).}
    \label{tab:CrossValTabl}
    \begin{tablenotes}
    \item[*] Contains Baseline Uncorrected Data
    \item[**] Contains Increased Noise
    \end{tablenotes}
    \end{threeparttable}
\end{table}

\pagebreak
\subsection{Convolutional Neural Network Results}
Current literature utilizes data augmentation techniques that are computationally expensive and require large data sets (>1000 spectra), such as Generative Adversarial Networks.\cite{WuWan2021} While these methods are extremely effective, the requirement of a large dataset makes such techniques impractical for concentration data. It is for this reason that the augmentation strategy used in this work is based upon the spectral peak distribution of the dataset and involves simple transformations of existing data. More advanced augmentation methods, or even a more detailed study into optimizing this augmentation strategy would be required to ensure greater reliability of the algorithm.

The Convolutional Neural Network was evaluated using average cross-validation accuracy score as well as the average \acrshort{cv} validation loss. Prediction accuracy of the \acrshort{cnn} model was heavily reliant on data quality with the \acrshort{r6g} dataset producing the best results ($85.2\% \pm 4.4\%$). Despite the limitations of the triclosan and chlorpyrifos datasets in terms of reduced size, increased noise, and high variance at low concentrations, prediction accuracy is still considerable ($82.7\% \pm 6.9\%$, and $70\% \pm 12.6\%$ respectively) with usage of the best treatment. In this work, categorical accuracy is used as an accuracy metric due to the CNN model being purely exploratory. A optimized model should use more refined accuracy parameters such as f1 score or area under receiver operating characteristic curve, depending on intended application. Additionally, most applications should incorporate a euclidean distance (least squares type) or threshold based metric when training. Euclidean distance metrics will make more close predictions at a cost to number of perfect predictions, while threshold based metrics will prioritize over/under predicting or staying within a set range. Scores for an example regression loss function are included in the supplemental information.

\begin{figure}[H]
\includegraphics[scale = 0.625]{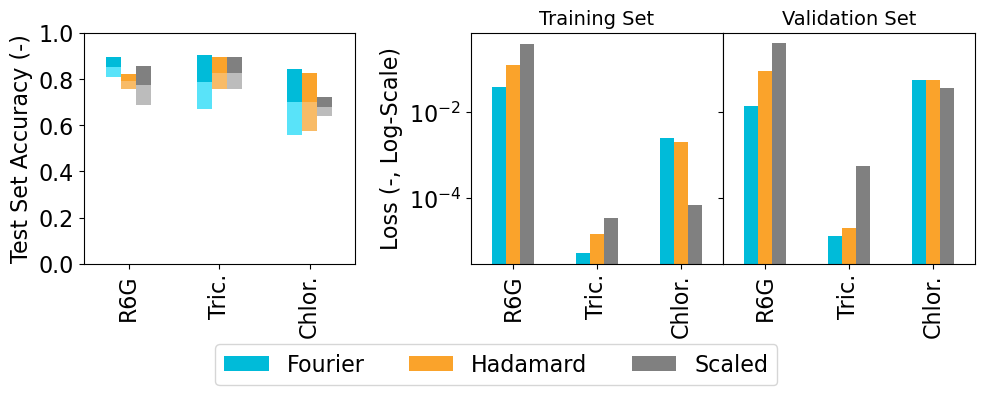}
\caption{Average 5-fold cross-validation results for Convolutional Neural Network. Test set average accuracy ratios plotted with standard deviation. Training and validation average losses plotted in log scale.}
\label{fgr:CNNAcc}
\end{figure}

Figure \ref{fgr:CNNAcc} displays the testing set accuracy of the three models for each of the three chemicals. In the case of \acrshort{r6g}, the Fourier transform performs best, with Hadamard having a slight advantage over scaled. For the triclosan dataset scaled and Hadamard perform equally while Fourier efficacy is reduced. Finally, for the chlorpyrifos dataset both the transformed sets outperformed the scaled set, with the Hadamard being preferred due to its lower standard deviation. Overall, the Hadamard transform shows equal or better performance when compared to the scaled set. The Fourier transform shows extremely good performance on \acrshort{r6g}, but may be less reliable than the Hadamard since it underperforms simple scaling on triclosan. Both frequency domain transforms are more susceptible to poor test set selection than the standard scaled model. This can be seen as increased standard deviation in these models when applied to triclosan and chlorpyrifos, as compared to the \acrshort{r6g} values. Increasing dataset size will likely help the transformed models outperform the scaled models.

The training and validation losses are also presented in Figure \ref{fgr:CNNAcc} on a log scale. The validity of observations from test set results are limited due to generalizability concerns. As such consideration should be paid to training and validation losses as the model selection metric. Training and validation losses are very similar for \acrshort{r6g}, suggesting minimal to no over or underfit. For triclosan and chlorpyrifos there is some overfit for all models. Transformed model losses are lower than scaled, except in chlorpyrifos, with Fourier slightly outperforming Hadamard. Training and validation accuracies for the transformed models are generally higher than scaled. The mismatch between train/validation performance and test performance suggests that transformed models are better than scaled models but are limited by the data augmentation strategy. Despite the data augmentation strategy having significant room for improvement, the test scores are high given the type of problem (multiclass classification), suggesting that the augmentation is not poorly conceived. The test results of \acrshort{r6g} best match the corresponding train/val results, further supporting the idea that dataset size hampers Fourier and Hadamard performance on triclosan and chlorpyrifos.

\pagebreak
\subsection{Identification of Characteristic Peaks}
A key factor in the analysis of Raman spectra is the characterization of a chemical in terms of its key peaks. These peaks are associated with various chemical bonds present in the compound. As the characteristics peaks are directly linked to the chemical structure, full knowledge of the peak locations can uniquely identify the compound being analyzed. Tracking the shift and intensification of key peaks enables many analytical techniques such as measurement of species via functionalized surfaces\cite{XuSon2021}, reaction extent monitoring\cite{SveJos1999}, and measurement of changes in polymer deformation/orientation.\cite{LyoChr1998}

In a similar fashion, when training a machine learning model with complete spectra, the model will rely on the data at certain wavenumbers more than others. As each wavenumber is a feature in our models, this is represented by the feature importances of the trained model. These feature importances have an unique relationship with the characteristic peaks of the spectra. Key peaks will generally be more sensitive to increased concentration of the analyte, and will therefore have an elevated importance. This allows for peaks in the spectra to be identified as potential characteristic peaks based on their importance scores. This can provide greater insight about the collected spectra by identifying peaks that are more sensitive or insensitive to variation in concentration. Additionally, this could have some potential application for the identification of key peaks in unknown or convoluted spectra. 

\begin{figure}[H]
\includegraphics[scale=0.5]{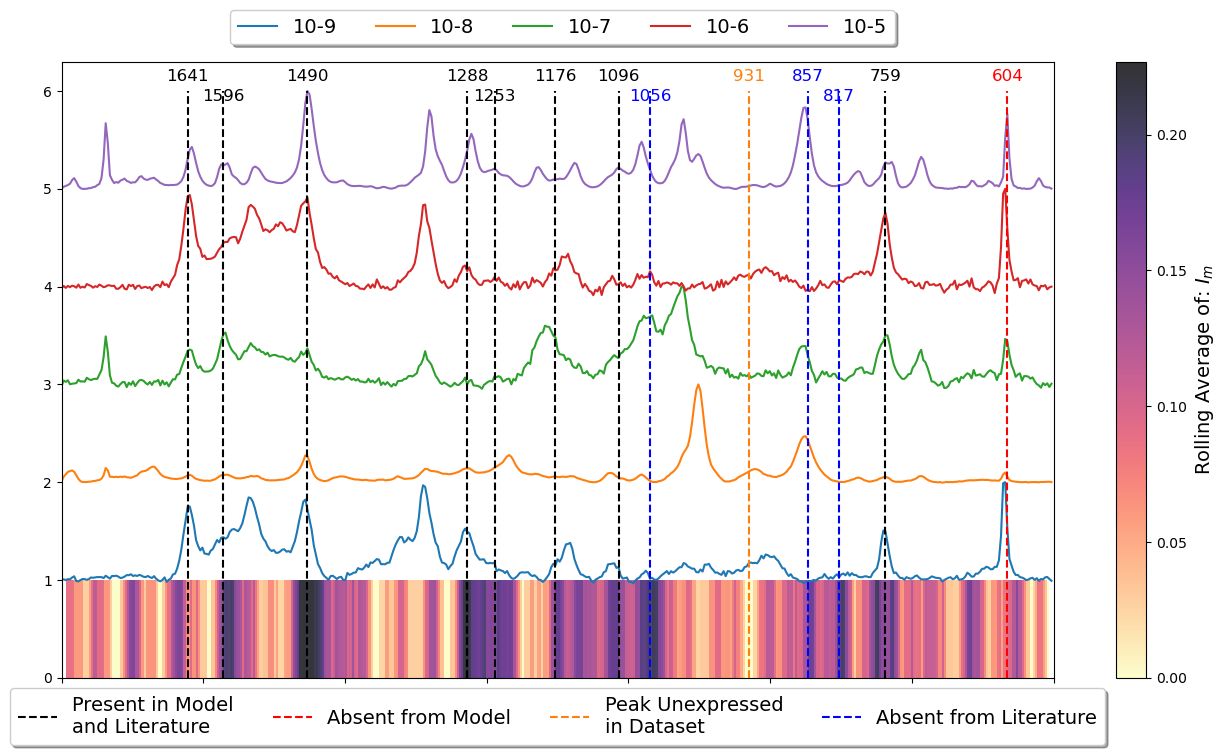}
\caption{Characteristic Raman peaks of R6G (AgNano rings method) in comparison with peaks used by random forest for classification. Normalized average spectra at each concentration in dataset (offset by class). Colorbar corresponds to the rolling average (two neighboring wavenumbers in each direction) of the modified Importance Score ($I_m$).}
\label{fgr:R6gCharP}
\end{figure}

Figure \ref{fgr:R6gCharP} shows the normalized average Raman spectra of each concentration for R6G. Similar spectra for triclosan and chlorpyrifos are shown in Figures S3 and S4. The colourbar and respective colourmap at the bottom of the figure is a measure of how important each wavenumber is to the random forest algorithm. There is some peak shift for each chemical when comparing the spectra from these datasets to that of literature due to variations in \acrshort{sers} substrate and analytical techniques. These peak shifts are typically less than 20 wavenumbers with most being less than 10, within expectation for \acrshort{sers} spectra.\cite{HilSto1984} Tables S1-3 show detailed assignments of each peak identified in literature and the model as well as the potential shift between the data and literature.

In the figures, peaks of high algorithmic importance that are identified in literature reference, or the bulk spectra reference in the case of triclosan, and are marked and labeled in black. Peaks that are identified in the reference but are not considered important by the algorithm are marked and labelled in red. Peaks that are identified in the reference but are not seen clearly in the dataset are marked and labelled in orange. Finally, peaks with considerable algorithmic importance that are not identified by the reference are marked in blue. These can be unidentified so far, representative of some common bonds that cannot be considered characteristic, or related to the noise/baseline in the spectra. 

The importance of each wavenumber is calculated via impurity method. With spectral data the number of features(each wavenumber) is very high. Therefore, many feature importances will be zero or near zero. Since importance scores across all features will add to one, the existence of these numerous near zero values can dampen the importance of the major peaks used by the algorithm, making the interpretation of results difficult. To solve this issue, a transformation function, described by Equation \ref{eqn:ModifiedImportance}, is applied to the importance scores. This has the effect of making the important areas of the spectra more noticeable and the result value is called the modified importance score ($I_m$). Peak identification plots display a rolling average (two neighbors in either direction) of modified importance score to improve readability.

\begin{equation}
I_m = |\frac{1}{ln(I)}|
\label{eqn:ModifiedImportance}
\end{equation}

Labels of characteristic peaks are taken from literature for both \acrshort{r6g} and chlorpyrifos.\cite{MaLuy2020,HilSto1984} For triclosan, four bulk triclosan powder spectra were collected for comparison.\cite{KanWu2023} These bulk powder spectra are averaged and presented in Figure S3 as the 'Raw' spectra. Due to the peak enhancing and possible peak shift of \acrshort{sers}, the matching of the \acrshort{sers} data to bulk powder data in the triclosan case will not be as good as the matching for the other chemicals, which use \acrshort{sers} references. 

There are important caveats that need to be discussed when matching characteristic peaks based on importance score. Firstly, the algorithm may assign high importance the entire width of the peak, or a particular representative wavenumber at some point in the peak. In the \acrshort{r6g} spectra (Figure \ref{fgr:R6gCharP}), the high importance peak at 1641 $cm^{-1}$, corresponding to aromatic C-C stretch, is identified by a band at the middle of its positive incline. In contrast the 1490 $cm^{-1}$ is of high importance across its entirety. Secondly, importance will only be assigned to as many peaks as needed. In the chlorpyrifos spectra (Figure S4), classification can be achieved largely by considering the only the P=S stretch at 624 $cm^{-1}$ therefore minimal importance is assigned to other peaks and the entirety of the P=S stretch has extremely high importance. Finally, the importance of some peaks may instead be assigned to a different peak that is well correlated to it. As an example, C-C-C ring in plane bend at 604 $cm^{-1}$ in the \acrshort{r6g} spectra is unused by the algorithm. However, it has a high correlation to the peak near 760$ cm^{-1}$ (>0.8 Spearman correlation), which is used by the algorithm and has a high importance. This means that using the 604 $cm^{-1}$ peak provides information that is already known from the 760 $cm^{-1}$ peak and is therefore considered less important.

Despite these limitations, algorithmic importance identifies most characteristic peaks in both chlorpyrifos and \acrshort{r6g}, missing only 1 peak in each. All three species have some algorithmically important wavenumbers that do not have an assignment in the reference. The triclosan spectra were difficult to properly analyze as bulk reference samples have different peaks from a liquid sample \acrshort{sers} spectra. Even so, the most crucial peaks at 782$cm^{-1}$ and 703$cm^{-1}$, which are used by the original researchers, are well identified.\cite{KanWu2023}

\pagebreak
\section{Conclusions} 
This work applied machine classification techniques to surface-enhanced Raman spectroscopy data with the intent of determining concentration. Firstly, through standard machine learning techniques medium-high prediction accuracies (>80\%) are achievable even using uncorrected, unfiltered, mixed origin datasets. Next, using convolutional neural networks with a data augmentation strategy based on simple transformations of data, with a sufficiently sized, moderately clean rhodamine 6G dataset(>100 spectra) prediction accuracies of above 85\% were achieved via the Fourier transform. For two smaller datasets with lower quality, triclosan and chlorpyrifos, prediction accuracies of 82\% and 70\%, respectively, were achieved. Both the Fourier and Hadamard transforms are shown to be useful tools in improving prediction accuracy, with the Hadamard performing especially well across datasets in standard and the CNN models. Further tuning \acrshort{cnn} architecture and augmentation strategy could provide more promising results. Finally, machine learning models for concentration prediction have good matching with literature assignment of characteristic peaks and have potential as a tool for identification of characteristic peaks when they are unknown. Further refinement of \acrshort{sers} as a concentration detection technique via machine learning has potential to allow for in-field measurement of trace organic contaminants at levels previously impractical.
\pagebreak

\pagebreak

\begin{acknowledgement}

The authors thank the MITACs organization which provided a portion of the funding for this international cooperation via the MITACs Globalink Research Award (RID: IT28380). J. B. You acknowledges support from the Korea National Research Foundation of Korea (NRF) grant funded by the Korea government (MSIT) (No. NRF-2021R1F1A1045909). The project is partly supported by Discovery project and Alliance Grant from the Natural Science and Engineering Research Council of Canada (NSERC), and by Advanced Program from Alberta Innovates. Computations were performed on the Niagara supercomputer at the SciNet HPC Consortium. SciNet is funded by: the Canada Foundation for Innovation; the Government of Ontario; Ontario Research Fund - Research Excellence; and the University of Toronto.

\end{acknowledgement}
\begin{suppinfo}
Supporting tables and figures available at: \href{run:./Supporting.tex}{\textcolor{blue}{Supporting Information}}
\\
Code base available on GitHub at: \href{https://github.com/VishnuJay/Determination-of-Trace-Organic-Contaminant-Concentration-via-Machine-Classification-of-Raman-Spectra}{\textcolor{blue}{Code Repository}}
\\
Raw spectral data available on GitHub at: \href{https://github.com/VishnuJay/Determination-of-Trace-Organic-Contaminant-Concentration-via-Machine-Classification-of-Raman-Spectra/tree/main/Finalized%20Code/RAMAN%20Data%20MITACs}{\textcolor{blue}{Raw Spectral Data}}
\\
\end{suppinfo}
\bibliography{ref}

\end{document}


\begin{doublespace}

\maketitle

\setcounter{figure}{0}
\makeatletter 
\renewcommand{\thefigure}{S\arabic{figure}}
\renewcommand{\thetable}{S\arabic{table}}

\pagebreak
\section{Tables and Figures}
\begin{figure}[h]
\includegraphics[scale= 0.6]{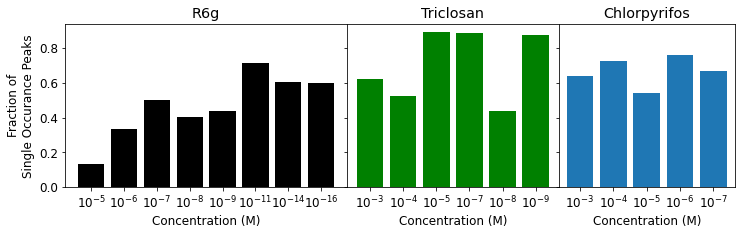}
\caption{Fraction of single occurrence peaks in each chemical spectral database.}
\label{fgr:SinglePeaks}
\end{figure}

\begin{figure}[H]
\includegraphics[scale=0.475]{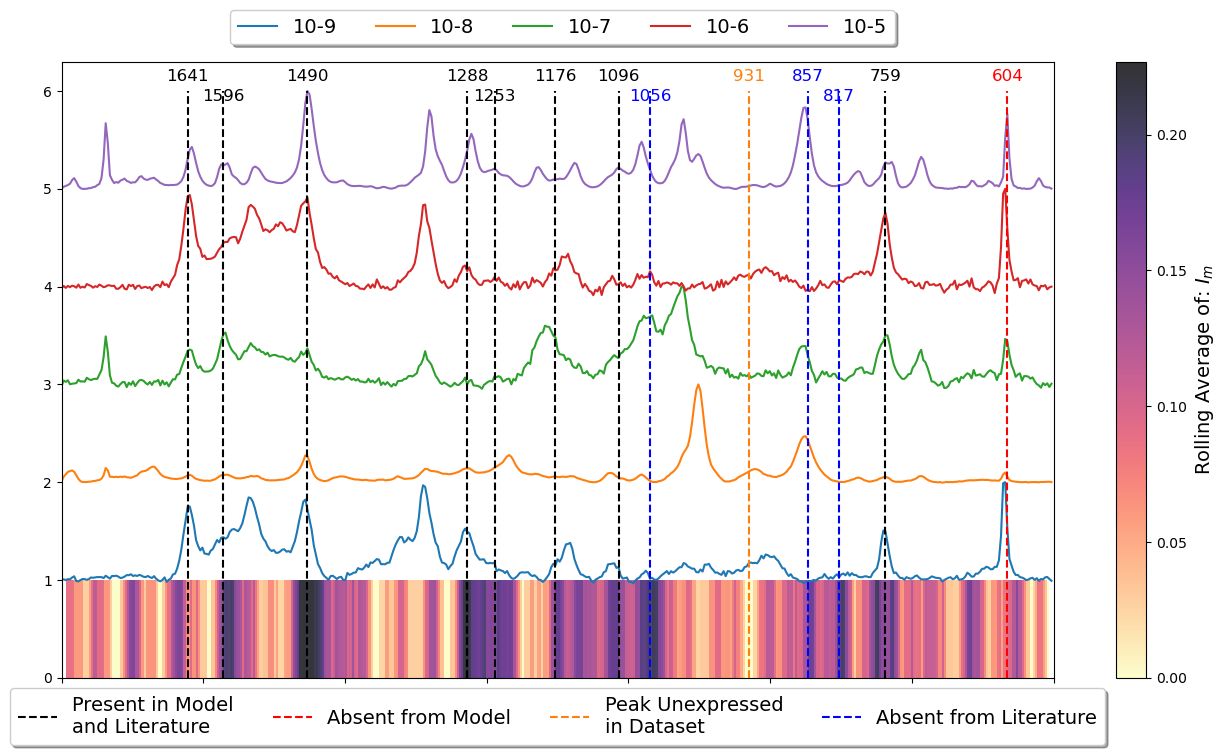}
\caption{Characteristic Raman peaks of R6G (AgNano rings method) in comparison with peaks used by random forest for classification.}
\label{fgr:R6gCharP}
\end{figure}

\begin{table}[H]
    \centering
    \begin{tabular}{|l|l|r|r|}
    \hline
        \multicolumn{4}{|c|}{\makecell[c]{\textbf{R6g}}}\\ \hline
        \makecell[c]{Wave Number\\(Model/Lit.(Shift))} & \makecell[c]{Literature\\Identification} & \makecell[c]{Raw\\Imp.(I)} & \makecell[c]{Rolling\\Average of $I_m$} \\ \hline
        \multicolumn{4}{|c|}{\makecell[c]{\textit{Represented in Data, Model and Literature}}} \\ \hline
        1641/1650(9) & Aromatic C-C Stretch & 0.012 & 0.165 \\ \hline
        1592/1597(5) & Aromatic C-C Stretch & 0.013 & 0.202 \\ \hline
        1490/1509(19) & Aromatic C-C Stretch & 0.016 & 0.222 \\ \hline
        1288/1310(22) & Aromatic C-C Stretch & 0.017 & 0.226 \\ \hline
        1253/1268(15) & C-O-C Stretch & 0.008 & 0.190 \\ \hline
        1176/1183(7) & C-H In Plane Bend & 0.009 & 0.179 \\ \hline
        1096/1088(8) & C-H In Plane Bend & 0.006 & 0.181 \\ \hline
        759/776(17) & C-H Out of Plane Bend & 0.016 & 0.205 \\ \hline
        \multicolumn{4}{|c|}{\makecell[c]{\textit{Represented in Data and Literature, Ignored by Model}}} \\ \hline
        604/614(10) & C-C-C Ring in Plane Bend & - & - \\ \hline
        \multicolumn{4}{|c|}{\makecell[c]{\textit{Represented in Literature, Unexpressed in Data}}} \\ \hline
        931 & C-H Out of Plane Bend & - & - \\ \hline
        \multicolumn{4}{|c|}{\makecell[c]{\textit{Represented in Model/Data, Unidentified in Literature}}} \\ \hline
        1056 & - & 0.014 & 0.207 \\ \hline
        857 & - & 0.004 & 0.174 \\ \hline
        806 & - & 0.012 & 0.201 \\ \hline
    \end{tabular}
        
    \caption{Rhodamine 6G peak assignment from literature compared to importance.}
    \label{tab:R6gPeakAss}
\end{table}

\begin{figure}[H]
\includegraphics[scale=0.475]{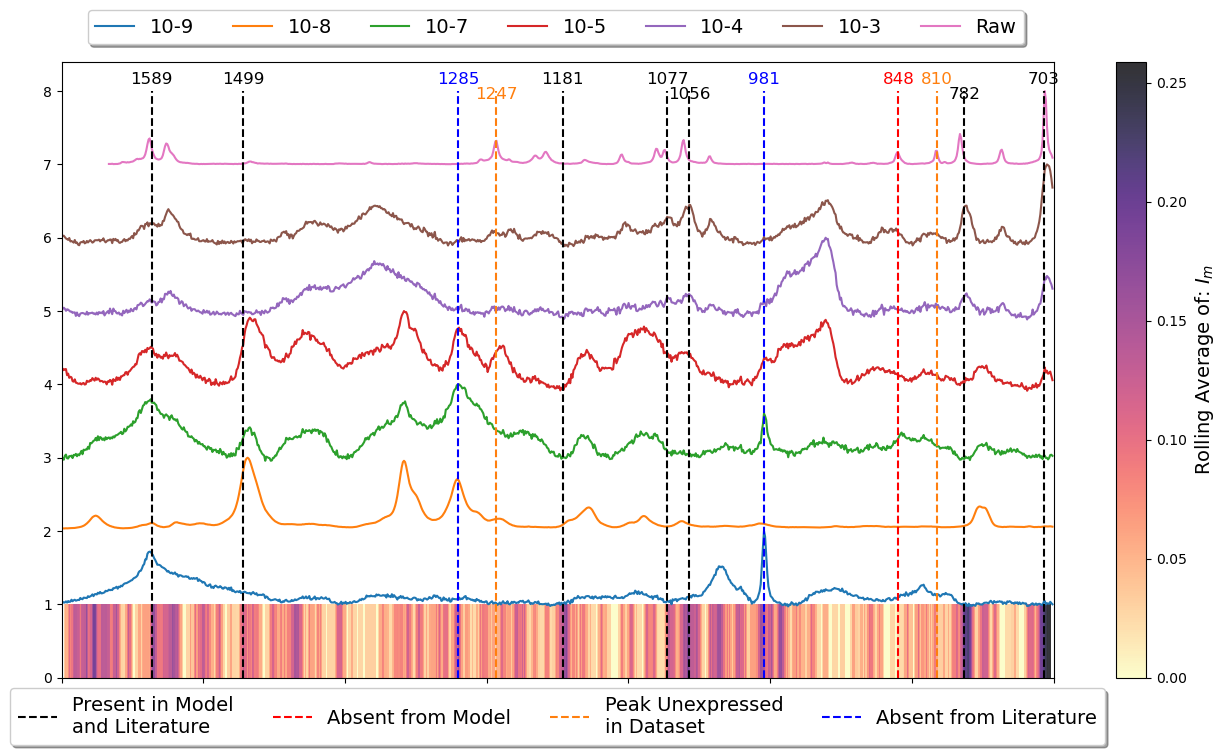}
\caption{Characteristic Raman peaks of triclosan in comparison with peaks used by random forest for classification.}
\label{fgr:TriCharP}
\end{figure}

\begin{table}[H]
    \centering
    \begin{tabular}{|c|r|r||c|r|r|}
    \hline
        \multicolumn{6}{|c|}{\makecell[c]{\textbf{Tric.}}} \\ \hline
        \makecell[c]{WaveNumber\\(Model)} & \makecell[c]{Raw\\Imp.(I)} & \makecell[c]{Rolling\\Average of $I_m$}  & \makecell[c]{WaveNumber\\(Model)} & \makecell[c]{Raw\\Imp.(I)} & \makecell[c]{Rolling\\Average of $I_m$} \\ \hline
        \multicolumn{3}{|c||}{\makecell[c]{\textit{Represented in Data, Model and Bulk}}} & \multicolumn{3}{c|}{\makecell[c]{\textit{Represented in Data and Bulk, Ignored by Model}}} \\ \hline
        1589 & 0.004 & 0.167 & 848 & - & - \\ \hline
        1499 & 0.005 & 0.135 &  \multicolumn{3}{c|}{\makecell[c]{\textit{Represented in Bulk, Unexpressed in Data}}} \\ \hline
        1181 & 0.004 & 0.163 & 1247 & - & -\\ \hline
        1077 & 0.009 & 0.137 & 810 & - & -\\ \hline
        1056 & 0.011 & 0.182 & \multicolumn{3}{c|}{\makecell[c]{\textit{Represented in Model, Unexpressed in Bulk}}} \\ \hline
        782 & 0.021 & 0.226 & 1285 & 0.003 & 0.155 \\ \hline
        703 & 0.026 & 0.257 & 981 & 0.010 & 0.158 \\ \hline
        \end{tabular}
    \caption{Triclosan peak assignment from bulk powder spectra compared to importance.}
    \label{tab:TriPeakAss}
\end{table}

\begin{figure}[H]
\includegraphics[scale=0.475]{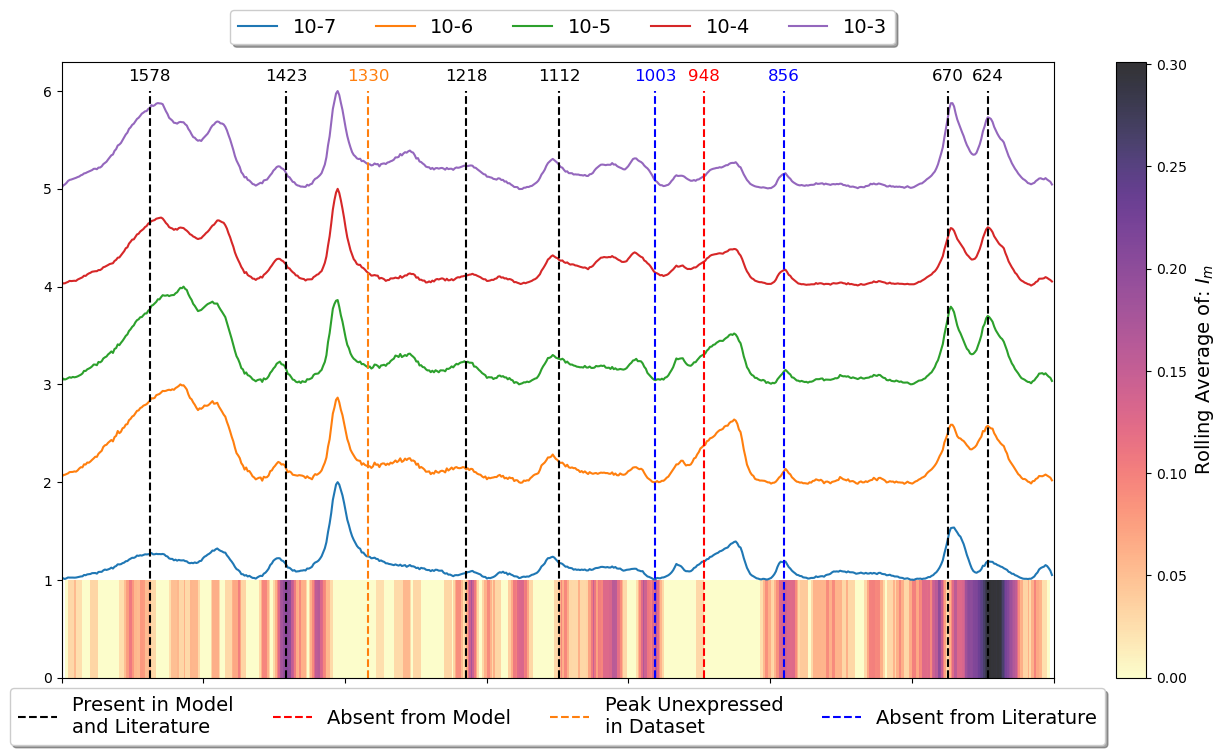}
\caption{Characteristic Raman peaks of chlorpyrifos in comparison with peaks used by random forest for classification.}
\label{fgr:ChlorCharP}
\end{figure}

\begin{table}[H]
    \centering
    \begin{tabular}{|l|l|r|r|}
    \hline
        \multicolumn{4}{|c|}{\makecell[c]{\textbf{Chlor.}}}\\ \hline
        \makecell[c]{Wave Number\\(Model/Lit.(Shift))} & \makecell[c]{Literature\\Identification} & \makecell[c]{Raw\\Imp.(I)} & \makecell[c]{Rolling\\Average of $I_m$} \\ \hline
        \multicolumn{4}{|c|}{\makecell[c]{\textit{Represented in Data, Model and Literature}}}\\ \hline
        1578/1571(7) & Ring Stretching & 0.001 & 0.081 \\ \hline
        1423/1439(16) & Cl Ring & 0.011 & 0.200 \\ \hline
        1218/1210(8) & Cl Ring Vibration & 0.007 & 0.177 \\ \hline
        1112/1092(20) & Cl Ring Wagging & 0.002 & 0.093 \\ \hline
        670/685(15) & P=S Stretch & 0.008 & 0.190 \\ \hline
        624/601(23) & P=S Stretch & 0.040 & 0.301 \\ \hline
        \multicolumn{4}{|c|}{\makecell[c]{\textit{Represented in Data and Literature, Ignored by Model}}} \\ \hline
        948/970(22) & Cl Ring Wagging & - & - \\ \hline
        \multicolumn{4}{|c|}{\makecell[c]{\textit{Represented in Literature, Unexpressed in Data}}} \\ \hline
        1331 & Cl Ring Vibration & - & - \\ \hline
        \multicolumn{4}{|c|}{\makecell[c]{\textit{Represented in Model/Data, Unidentified in Literature}}} \\ \hline
        1003 & - & 0.008 & 0.137 \\ \hline
        856 & - & 0.002 & 0.155 \\ \hline
        \end{tabular}
    \caption{Chlorpyrifos peak assignment from literature compared to importance.}
    \label{tab:ChlorPeakAss}
\end{table}

\begin{figure}[H]
\includegraphics[scale=0.5]{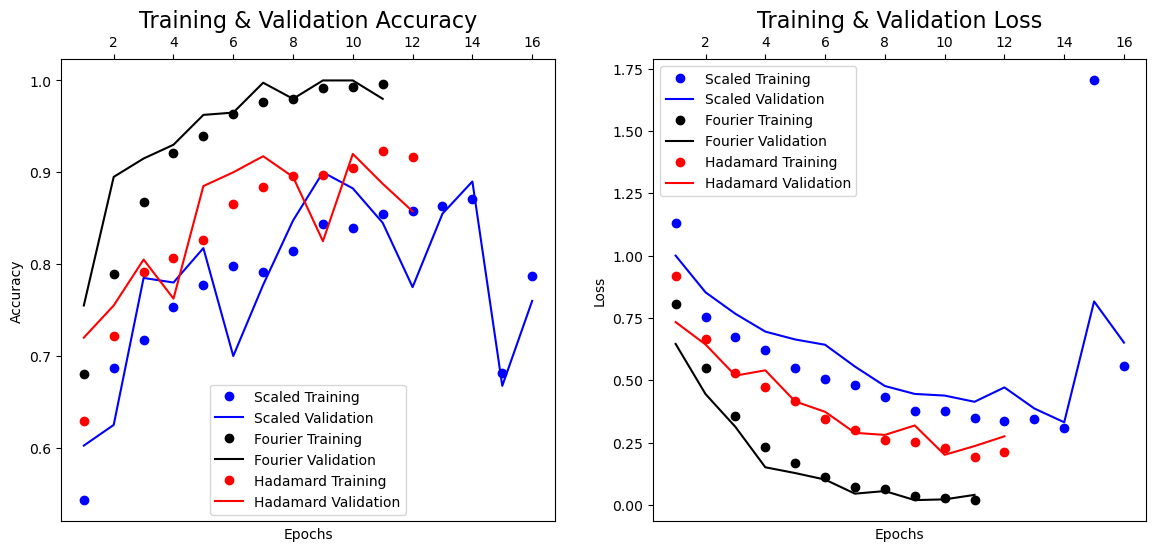}
\caption{Example learning curve for the training of the R6G CNN for each transformation.}
\label{fgr:R6gLearnCurve}
\end{figure}

\begin{figure}[H]
\includegraphics[scale=0.5]{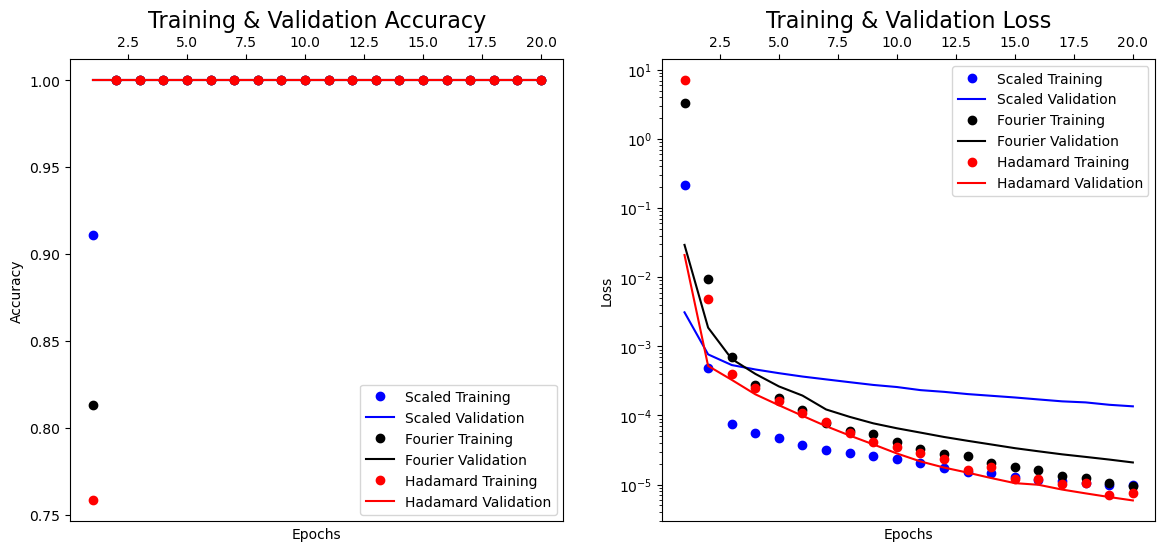}
\caption{Example learning curve for the training of the triclosan CNN for each transformation. (Accuracy curve affected by low sample size.)}
\label{fgr:TriLearnCurve}
\end{figure}

\begin{figure}[H]
\includegraphics[scale=0.5]{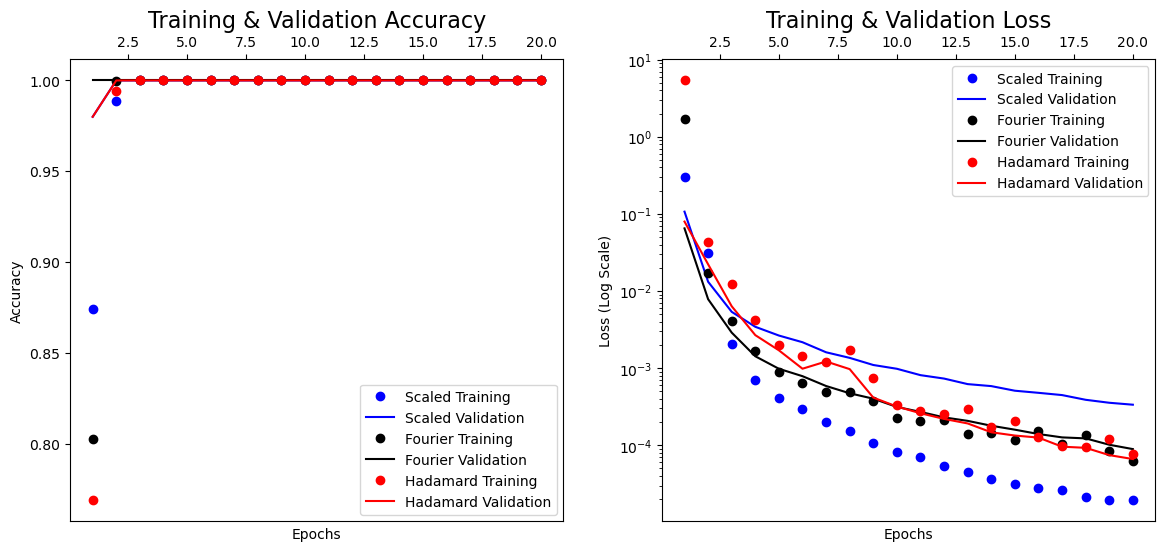}
\caption{Example learning curve for the training of the chlorpyrifos CNN for each transformation. (Accuracy curve affected by low sample size.)}
\label{fgr:ChlorLearnCurve}
\end{figure}

\begin{figure}[H]
\includegraphics[scale=0.4]{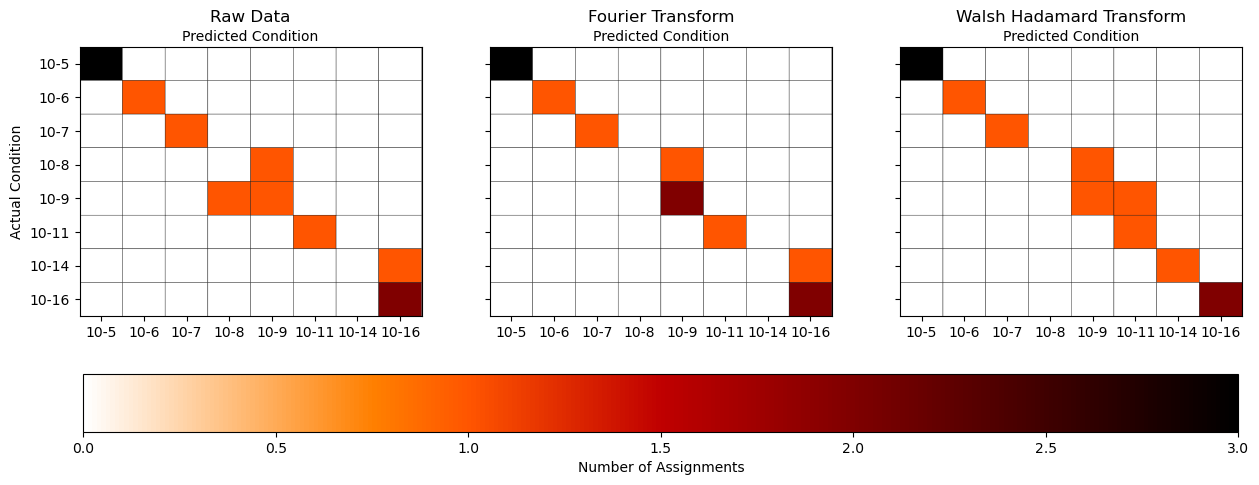}
\caption{Example confusion matrix for a particular random seed train-test split result of the R6G models.}
\label{fgr:Confusion}
\end{figure}

\section{CNN Cross Validation Raw Data}

\begin{equation}
\begin{aligned}
E_{reg} = \sum_{i=0}^{n_p}{\frac{(y_{pred}(i) - y_{true}(i))^2}{n_c^2}}
\label{eqn:RegErr}
\end{aligned}
\end{equation}
\\
$E_{reg}$ is the Regression Error.
\\
$y_{pred}(i) - y_{true}(i)$ is the categorical distance between the $i^{th}$ prediction and the $i^{th}$ true value.
\\
$n_c$ is the number of classification categories.
\\
$n_p$ is the number or predictions made.
\pagebreak
\subsection{RHODAMINE 6G}
\begin{verbatim}
SCALED											
Test	Acc	Test	Loss	Train	Acc	Train	Loss	Val_Acc	Val_Loss	Regression	Error
0.696	0.517	0.865	0.326	0.882	0.4	0.344					
FOURIER											
Test	Acc	Test	Loss	Train	Acc	Train	Loss	Val_Acc	Val_Loss	Regression	Error
0.783	0.882	1	0.003	1	0.004	0.25					
HADAMARD											
Test	Acc	Test	Loss	Train	Acc	Train	Loss	Val_Acc	Val_Loss	Regression	Error
0.783	0.407	0.952	0.126	0.98	0.075	0.125					
----------------------------											
SCALED											
Test	Acc	Test	Loss	Train	Acc	Train	Loss	Val_Acc	Val_Loss	Regression	Error
0.696	0.538	0.857	0.34	0.863	0.333	0.469					
FOURIER											
Test	Acc	Test	Loss	Train	Acc	Train	Loss	Val_Acc	Val_Loss	Regression	Error
0.87	0.248	0.994	0.022	1	0.009	0.172					
HADAMARD											
Test	Acc	Test	Loss	Train	Acc	Train	Loss	Val_Acc	Val_Loss	Regression	Error
0.826	0.328	0.952	0.137	0.973	0.097	0.313					
----------------------------											
SCALED											
Test	Acc	Test	Loss	Train	Acc	Train	Loss	Val_Acc	Val_Loss	Regression	Error
0.913	0.467	0.863	0.315	0.84	0.373	0.031					
FOURIER											
Test	Acc	Test	Loss	Train	Acc	Train	Loss	Val_Acc	Val_Loss	Regression	Error
0.87	0.431	0.986	0.048	1	0.005	0.141					
HADAMARD											
Test	Acc	Test	Loss	Train	Acc	Train	Loss	Val_Acc	Val_Loss	Regression	Error
0.826	0.769	0.951	0.133	0.955	0.098	0.391					
----------------------------											
SCALED											
Test	Acc	Test	Loss	Train	Acc	Train	Loss	Val_Acc	Val_Loss	Regression	Error
0.826	0.462	0.918	0.197	0.918	0.213	0.234					
FOURIER											
Test	Acc	Test	Loss	Train	Acc	Train	Loss	Val_Acc	Val_Loss	Regression	Error
0.826	0.598	0.973	0.092	1	0.005	0.313					
HADAMARD											
Test	Acc	Test	Loss	Train	Acc	Train	Loss	Val_Acc	Val_Loss	Regression	Error
0.739	0.617	0.957	0.129	0.973	0.094	0.5					
----------------------------											
SCALED											
Test	Acc	Test	Loss	Train	Acc	Train	Loss	Val_Acc	Val_Loss	Regression	Error
0.739	1.716	0.723	0.698	0.77	0.646	0.344					
FOURIER											
Test	Acc	Test	Loss	Train	Acc	Train	Loss	Val_Acc	Val_Loss	Regression	Error
0.913	0.338	0.991	0.031	0.97	0.047	0.203					
HADAMARD											
Test	Acc	Test	Loss	Train	Acc	Train	Loss	Val_Acc	Val_Loss	Regression	Error
0.783	0.371	0.958	0.11	0.985	0.073	0.328					
----------------------------		



*********************************************************											
SCALED	AVERAGE										
Test	Acc	Test	Loss	Train	Acc	Train	Loss	Val_Acc	Val_Loss	Regression	Error
0.774	0.74	0.845	0.375	0.854	0.393	0.284					
SCALED	STANDARD	DEVIATION									
Test	Acc	Test	Loss	Train	Acc	Train	Loss	Val_Acc	Val_Loss	Regression	Error
0.084	0.489	0.065	0.169	0.049	0.142	0.147					
----------------------------											
FOURIER	AVERAGE										
Test	Acc	Test	Loss	Train	Acc	Train	Loss	Val_Acc	Val_Loss	Regression	Error
0.852	0.5	0.989	0.039	0.994	0.014	0.284					
FOURIER	STANDARD	DEVIATION									
Test	Acc	Test	Loss	Train	Acc	Train	Loss	Val_Acc	Val_Loss	Regression	Error
0.044	0.224	0.009	0.03	0.012	0.017	0.147					
----------------------------											
HADAMARD	AVERAGE										
Test	Acc	Test	Loss	Train	Acc	Train	Loss	Val_Acc	Val_Loss	Regression	Error
0.791	0.498	0.954	0.127	0.973	0.087	0.284					
HADAMARD	STANDARD	DEVIATION									
Test	Acc	Test	Loss	Train	Acc	Train	Loss	Val_Acc	Val_Loss	Regression	Error
0.033	0.168	0.003	0.009	0.01	0.011	0.147					
\end{verbatim}
\pagebreak
\subsection{TRICLOSAN}
\begin{verbatim}
SCALED											
Test	Acc	Test	Loss	Train	Acc	Train	Loss	Val_Acc	Val_Loss	Regression	Error
0.909	0.354	1	0	1	0.001	0.25					
Fourier											
Test	Acc	Test	Loss	Train	Acc	Train	Loss	Val_Acc	Val_Loss	Regression	Error
0.909	0.178	1	0	1	0	0.25					
HADAMARD											
Test	Acc	Test	Loss	Train	Acc	Train	Loss	Val_Acc	Val_Loss	Regression	Error
0.909	0.323	1	0	1	0	0.25					
----------------------------											
SCALED											
Test	Acc	Test	Loss	Train	Acc	Train	Loss	Val_Acc	Val_Loss	Regression	Error
0.727	1.073	1	0	1	0.001	0.167					
Fourier											
Test	Acc	Test	Loss	Train	Acc	Train	Loss	Val_Acc	Val_Loss	Regression	Error
0.727	1.798	1	0	1	0	0.167					
HADAMARD											
Test	Acc	Test	Loss	Train	Acc	Train	Loss	Val_Acc	Val_Loss	Regression	Error
0.727	1.409	1	0	1	0	0.167					
----------------------------											
SCALED											
Test	Acc	Test	Loss	Train	Acc	Train	Loss	Val_Acc	Val_Loss	Regression	Error
0.8	0.698	1	0	1	0	0.139					
Fourier											
Test	Acc	Test	Loss	Train	Acc	Train	Loss	Val_Acc	Val_Loss	Regression	Error
0.6	70.66	1	0	1	0	0.194					
HADAMARD											
Test	Acc	Test	Loss	Train	Acc	Train	Loss	Val_Acc	Val_Loss	Regression	Error
0.8	30.443	1	0	1	0	0.278					
----------------------------											
SCALED											
Test	Acc	Test	Loss	Train	Acc	Train	Loss	Val_Acc	Val_Loss	Regression	Error
0.8	1.825	1	0	1	0.001	0.222					
Fourier											
Test	Acc	Test	Loss	Train	Acc	Train	Loss	Val_Acc	Val_Loss	Regression	Error
0.8	0.692	1	0	1	0	0.139					
HADAMARD											
Test	Acc	Test	Loss	Train	Acc	Train	Loss	Val_Acc	Val_Loss	Regression	Error
0.8	1.452	1	0	1	0	0.139					
----------------------------											
SCALED											
Test	Acc	Test	Loss	Train	Acc	Train	Loss	Val_Acc	Val_Loss	Regression	Error
0.9	0.932	1	0	1	0.001	0.028					
Fourier											
Test	Acc	Test	Loss	Train	Acc	Train	Loss	Val_Acc	Val_Loss	Regression	Error
0.9	0.736	1	0	1	0	0.028					
HADAMARD											
Test	Acc	Test	Loss	Train	Acc	Train	Loss	Val_Acc	Val_Loss	Regression	Error
0.9	0.958	1	0	1	0	0.028					
----------------------------



*********************************************************											
SCALED	AVERAGE										
Test	Acc	Test	Loss	Train	Acc	Train	Loss	Val_Acc	Val_Loss	Regression	Error
0.827	0.976	1	0	1	0.001	0.161					
SCALED	STANDARD	DEVIATION									
Test	Acc	Test	Loss	Train	Acc	Train	Loss	Val_Acc	Val_Loss	Regression	Error
0.069	0.489	0	0	0	0	0.077					
----------------------------											
FOURIER	AVERAGE										
Test	Acc	Test	Loss	Train	Acc	Train	Loss	Val_Acc	Val_Loss	Regression	Error
0.787	14.813	1	0	1	0	0.161					
FOURIER	STANDARD	DEVIATION									
Test	Acc	Test	Loss	Train	Acc	Train	Loss	Val_Acc	Val_Loss	Regression	Error
0.115	27.929	0	0	0	0	0.077					
----------------------------											
HADAMARD	AVERAGE										
Test	Acc	Test	Loss	Train	Acc	Train	Loss	Val_Acc	Val_Loss	Regression	Error
0.827	6.917	1	0	1	0	0.161					
HADAMARD	STANDARD	DEVIATION									
Test	Acc	Test	Loss	Train	Acc	Train	Loss	Val_Acc	Val_Loss	Regression	Error
0.069	11.77	0	0	0	0	0.077					
\end{verbatim}
\pagebreak
\subsection{CHLORPYRIFOS}
\begin{verbatim}
SCALED											
Test	Acc	Test	Loss	Train	Acc	Train	Loss	Val_Acc	Val_Loss	Regression	Error
0.700	1.239	1.000	0.000	1.000	0.001	0.240					
Fourier											
Test	Acc	Test	Loss	Train	Acc	Train	Loss	Val_Acc	Val_Loss	Regression	Error
0.500	3.476	0.998	0.002	0.985	0.031	0.640												
HADAMARD											
Test	Acc	Test	Loss	Train	Acc	Train	Loss	Val_Acc	Val_Loss	Regression	Error
0.600	1.832	1.000	0.001	0.990	0.061	0.480					
----------------------------											
SCALED											
Test	Acc	Test	Loss	Train	Acc	Train	Loss	Val_Acc	Val_Loss	Regression	Error
0.600	0.780	1.000	0.000	0.995	0.031	0.600					
Fourier											
Test	Acc	Test	Loss	Train	Acc	Train	Loss	Val_Acc	Val_Loss	Regression	Error
0.700	1.035	1.000	0.000	0.990	0.028	0.120												
HADAMARD											
Test	Acc	Test	Loss	Train	Acc	Train	Loss	Val_Acc	Val_Loss	Regression	Error
0.600	1.318	0.999	0.003	0.995	0.068	0.160					
----------------------------											
SCALED											
Test	Acc	Test	Loss	Train	Acc	Train	Loss	Val_Acc	Val_Loss	Regression	Error
0.700	1.680	1.000	0.000	1.000	0.001	0.120					
Fourier											
Test	Acc	Test	Loss	Train	Acc	Train	Loss	Val_Acc	Val_Loss	Regression	Error
0.600	1.726	1.000	0.001	0.995	0.009	0.280					.				
HADAMARD											
Test	Acc	Test	Loss	Train	Acc	Train	Loss	Val_Acc	Val_Loss	Regression	Error
0.600	1.125	1.000	0.001	0.995	0.022	0.160					
----------------------------											
SCALED											
Test	Acc	Test	Loss	Train	Acc	Train	Loss	Val_Acc	Val_Loss	Regression	Error
0.700	1.397	1.000	0.000	0.990	0.037	0.240					
Fourier											
Test	Acc	Test	Loss	Train	Acc	Train	Loss	Val_Acc	Val_Loss	Regression	Error
0.900	0.327	0.999	0.006	0.990	0.115	0.040												
HADAMARD											
Test	Acc	Test	Loss	Train	Acc	Train	Loss	Val_Acc	Val_Loss	Regression	Error
0.800	0.773	1.000	0.001	0.995	0.029	0.080					
----------------------------											
SCALED											
Test	Acc	Test	Loss	Train	Acc	Train	Loss	Val_Acc	Val_Loss	Regression	Error
0.700	0.705	1.000	0.000	0.990	0.108	0.240					
Fourier											
Test	Acc	Test	Loss	Train	Acc	Train	Loss	Val_Acc	Val_Loss	Regression	Error
0.800	0.257	0.999	0.002	0.985	0.095	0.080												
HADAMARD											
Test	Acc	Test	Loss	Train	Acc	Train	Loss	Val_Acc	Val_Loss	Regression	Error
0.900	0.396	0.999	0.004	0.990	0.091	0.040					
----------------------------



*********************************************************											
SCALED	AVERAGE										
Test	Acc	Test	Loss	Train	Acc	Train	Loss	Val_Acc	Val_Loss	Regression	Error
0.680	1.160	1.000	0.000	0.995	0.036	0.288					
SCALED	STANDARD	DEVIATION									
Test	Acc	Test	Loss	Train	Acc	Train	Loss	Val_Acc	Val_Loss	Regression	Error
0.040	0.370	0.000	0.000	0.004	0.039	0.163					
----------------------------											
FOURIER	AVERAGE										
Test	Acc	Test	Loss	Train	Acc	Train	Loss	Val_Acc	Val_Loss	Regression	Error
0.700	1.364	0.999	0.002	0.989	0.056	0.288					
FOURIER	STANDARD	DEVIATION									
Test	Acc	Test	Loss	Train	Acc	Train	Loss	Val_Acc	Val_Loss	Regression	Error
0.141	1.183	0.001	0.002	0.004	0.041	0.163					
----------------------------											
HADAMARD	AVERAGE										
Test	Acc	Test	Loss	Train	Acc	Train	Loss	Val_Acc	Val_Loss	Regression	Error
0.700	1.089	1.000	0.002	0.993	0.054	0.288					
HADAMARD	STANDARD	DEVIATION									
Test	Acc	Test	Loss	Train	Acc	Train	Loss	Val_Acc	Val_Loss	Regression	Error
0.126	0.487	0.000	0.001	0.002	0.026	0.163					
\end{verbatim}
\end{doublespace}